\RequirePackage{fix-cm}
\documentclass[smallcondensed]{svjour3}
\smartqed %flush right qed marks, e.g. at end of proof

\usepackage{graphicx}
\usepackage{amsmath}
\usepackage{amssymb}
\usepackage[utf8]{inputenc}
\usepackage[hidelinks]{hyperref}
\hypersetup{urlcolor=blue, colorlinks=true,citecolor=blue}
\usepackage{natbib}
\setcitestyle{authoryear,aysep={},open={(},close={)}}
\usepackage{caption}
\usepackage{multirow}
\usepackage{lineno}

\newcommand{\mlnew}[1]{{\textcolor{black}{#1}}}
\newcommand{\mlfeb}[1]{{\textcolor{black}{#1}}}
\newcommand{\er}[1]{{\textcolor{black}{#1}}}

\begin{document}

\title{Bayesian Anomaly Detection and Classification}

%\titlerunning{Short form of title} %if too long for running head

\author{Ethan Roberts\and
Bruce A. Bassett\and 
Michelle Lochner}

\authorrunning{E. Roberts, B. A. Bassett, M. Lochner} %if too long for running head

\institute{E. Roberts$^{1,2}$ \at
%first address \\
\email{rbreth001@myuct.ac.za}
\and
B. A. Bassett$^{1,2,3,4}$ \at
%second address \\
\email{bruce.a.bassett@gmail.com}
\and
M. Lochner$^{2,3}$ \at
%third address \\
\email{michelle@aims.ac.za}
\and
1. University of Cape Town, Rondebosch, Cape Town, 7700 \at
2. African Institute of Mathematical Sciences, Muizenburg, Cape Town, 7950 \at
3. South African Radio Astronomical Observatory, Observatory, Cape Town, 7295 \at
4. South African Astronimcal Observatory, Observatory, Cape Town, 7295
}

\date{Received: date / Accepted: date}

\maketitle

\begin{abstract}
Statistical uncertainties are rarely incorporated in machine learning algorithms, especially for anomaly detection. Here we present the Bayesian Anomaly Detection And Classification (BADAC) formalism, which provides a unified statistical approach to classification and anomaly detection within a hierarchical Bayesian framework. BADAC deals with uncertainties by marginalising over the unknown, true, value of the data. Using simulated data with Gaussian noise, BADAC is shown to be superior to standard algorithms in both classification and anomaly detection performance in the presence of uncertainties, though with significantly increased computational cost. Additionally, BADAC provides well-calibrated classification probabilities, valuable for use in scientific pipelines. We show that BADAC can work in online mode and is fairly robust to model errors, which can be diagnosed through model-selection methods. In addition it can perform unsupervised new class detection and can naturally be extended to search for anomalous subsets of data. BADAC is therefore ideal where computational cost is not a limiting factor and statistical rigour is important. We discuss approximations to speed up BADAC, such as the use of Gaussian processes, and finally introduce a new metric, the Rank-Weighted Score (RWS), that is particularly suited to evaluating the ability of algorithms to detect anomalies.

%Insert your abstract here. Include keywords, PACS and mathematical
%subject classification numbers as needed.

\end{abstract}

%%%%%%%%%%%%%%%%%%%%%%%%%%%%%%%%%%%%%%%%%%%%%%%%%%

%%%%%%%%%%%%%%%%% BODY OF PAPER %%%%%%%%%%%%%%%%%%

\section{Introduction}

In any fully rigorous or scientific analysis, uncertainties must be quantified and propagated through the full analysis pipeline. This is difficult to do with traditional machine learning algorithms that do not explicitly take into account uncertainties on the data or features. As machine learning is increasingly given authority for making more important and high-risk decisions, (e.g. in self-driving cars), and with the potential for adversarial attacks \citep{adversarial}, there is an increasing need for interpretable models and rigorous statistical uncertainties on machine learning predictions. 

In classification problems class labels are typically inferred through the use of a separation boundary that is learned from training data \citep{fawzi2017} and is based on a score combined with a threshold. This threshold is often arbitrary or learned as a hyperparameter to minimise some chosen loss function \citep{niculescu-mizil2005}. Any resulting class  ``probabilities" are systematically distorted in ways unique to the classification algorithm used and are not true probabilities, though in some cases these can be calibrated in a frequentist sense with more training data using e.g. isotonic regression or Platt scaling \citep{MLprobs}. 

However, particularly in the physical sciences, we desire an algorithm that automatically outputs unbiased, accurate probabilities, since knowing the probabilities of an object belonging to various classes is typically more useful than the class label alone. The classification process is often just one step in a multi-stage pipeline, and it is important to propagate class uncertainties through the additional steps in the analysis pipeline. This need is especially true in cases where the true class labels of the training data are noisy or subjective, or the training data are not representative of the test set. An example in astronomy is provided by the photometric classification of type Ia supernovae which are subsequently used for studies of dark energy. Hard label classification leads to contamination from non-Ia supernovae that leads to biases in dark energy properties while fully propagating class probabilities instead allows for unbiased results at the end of the pipeline \citep{beams1, beams2}. 

In this context Bayesian methods are ideal \citep{bayes_book}, as they have been proven optimal for classification for certain loss metrics, e.g. \cite{optimal}, and allow the option of both  supervised or unsupervised classification \citep{cheeseman1996}. In the context of astronomy, Bayesian techniques have been applied to classification of transient objects such as supernovae \citep{connolly2009}.  A common limitation in the classification of noisy data however, is that the classes in the training data are typically represented by a single template with zero variability (e.g. \cite{sako2011}). This allows straightforward Bayesian methods to be applied but does not apply if there is significant intraclass variability. Ignoring this intraclass variability also makes principled anomaly detection challenging: how unlikely is an example if one doesn't know the underlying distribution within a class? 

Here we address these limitations, constructing what we will argue is a natural, statistically robust supervised Bayesian method that can simultaneously be used for both anomaly detection and classification in the presence of measurement uncertainties on all data. Our method works directly with raw data, requiring no feature extraction, and requires minimal assumptions about the nature of the anomalies or classes.

We begin by describing the formalism in section \ref{sec:BHM}. We introduce a new metric optimised for anomaly detection, namely the Rank-Weighted Score (RWS) in section \ref{sec:RWS_score} (the other metrics we also use to assess algorithm performance are discussed in appendix \ref{sec:metrics}). Finally, we compare algorithm performance against various benchmark algorithms on simulated data in section \ref{sec:results}. The full derivation of the formalism is given in appendix \ref{sec:detailed-deriv}.

\section{The BADAC Formalism}\label{sec:BHM}

Bayesian Anomaly Detection and Classification (BADAC) is a Bayesian hierarchical formalism that provides the probabilities that a new measurement belongs to each of a set of known classes, and also ranks objects by the probability that they are anomalous. We make use of language common to machine learning by referring to {\it features} for the data for a specific instance of a class, {\it training data} (i.e. data for which we know the class label $\tau$) and referring to {\it test data} for data that we wish to classify as either anomalous or belonging to a known class $\tau$. 

We start by assuming we have a set of multiple classes, $\tau$, with each class having a set of (noisy) training feature instances $\{y_{o,j}^i\}_{\tau}$, along with associated uncertainties on the features, typically due to some form of measurement error. Here $i$ indexes the instances/examples in a class, while $j$ indexes the specific features within an instance.

\mlnew{For example, if we have a set of one-dimensional time-series data instances, then $j$ would index time. If our data were MNIST images\footnote{\url{http://yann.lecun.com/exdb/mnist/}}, then $j$ would index pixels, $\tau$ would index the integers $\{0,...,9\}$, and $i$ would index the examples of each digit. As implied by the above examples, the BADAC formalism we develop will work whether $j$ indexes a continuous underlying variable (e.g. time or space), or is a nominal index with no preferred ordering. The $o$ subscript denotes {\it observed} data, to distinguish it from the {\it true} (and unknown) underlying value which may differ because of noise or uncertainties. Allowing for the true and observed values separately will be useful, as we show later. For notational convenience we now suppress the class subscript on the $y_{o,j}^i$ but it should always be assumed. The test data to classify and rank for anomalousness, which we call $d$, should also be assumed to have measurement uncertainties which need to be included in any analysis. }

Since the final formulae are notationally complex, we gain intuition of the general case by considering the simplest possible example.  Assume each instance consists of just one feature, and that each class $\tau$ has only one instance. This would correspond to taking one point from one instance from each class in the time-series example. This simplification allows us to suppress both the $i$ and $j$ indices (as well as the $\tau$ label) and write $y_{o}$ for the moment. We assume $y_{o}$ has a known measurement error distribution. Then our final goal is to compute the posterior probability $P(\tau|d,y_o)$, of a single test data point $d$, with known error distribution, of belonging to a class $\tau$. This is discussed in more detail, along with the general derivation, in appendix \ref{sec:detailed-deriv} and shown schematically in figure \ref{fig:app:schem}. Bayes' theorem gives this posterior probability as:
\begin{equation}\label{eq:bayes}
P(\tau|d,y_o) \propto P(d,y_o|\tau) P(\tau)
\end{equation}
where $P(\tau)$ is the prior probability of belonging to class $\tau$.

If we have measurement errors associated with both $d$ and $y_o$, we cannot compute $P(\tau|d,y_o)$ with equation \ref{eq:bayes} as is. Instead we introduce a latent variable $y_t$ which represents the underlying true value, i.e. what would be measured if there were no measurement error. This can then be marginalised over to rigorously handle our uncertainty of the true value $y_t$. By application of the product rule for probability densities the posterior becomes:
\begin{align}
P(\tau|d,y_o) &\propto P(\tau) \int dy_t P(d,y_o,y_t|\tau)\\
 &\propto P(\tau) \int dy_t P(d,y_o|y_t,\tau) P(y_t|\tau)\\
 &\propto P(\tau) \int dy_t P(d|y_o,y_t,\tau) P(y_o|y_t,\tau) P(y_t|\tau) \label{eqn:product}
\end{align}
If we assume $d$ and $y_o$ are statistically independent of one another\footnote{We consider correlations in appendix \ref{sec:corr}.}, then $P(d|y_o,y_t,\tau) = P(d|y_t,\tau)$. We then arrive at: 
\begin{equation}\label{eq:bhm1}
P(\tau|d,y_o) \propto P(\tau) \int dy_t P(d|y_t,\tau) P(y_o|y_t,\tau) P(y_t|\tau)
\end{equation}
For Gaussian error distributions this can be solved analytically, as we show below.

Next let us generalise equation \ref{eq:bhm1} to allow each class $\tau$ to have multiple training instances, $\{y_o^i\}$, each still consisting of a single scalar feature value.  In this case we need to introduce $n$ latent variables, $y_t^i$, one for each instance in the class. Then the posterior probability that a new test instance $d$ belongs to class $\tau$ - assuming the instances in the training data are uncorrelated - is  given by (see Appendix \ref{sec:detailed-deriv} for derivation):
\begin{equation}\label{eq:generalcase1dp}
P(\tau|d,\{y_o^i\}) \propto P(\tau) \int dy_t^1 ... dy_t^n \left[\dfrac{1}{n} \sum_{i=1}^n P(d|y_t^i,\tau) \right] \prod_{i=1}^n P(y_o^i|y_t^i,\tau) \times \prod_{i=1}^n P(y_t^i|\tau)
\end{equation}
Here $P(d|y_t^i,\tau)$ is the likelihood of observing the data $d$, conditioned on both the class type and an estimate of the true values of the training data. $P(y_t^i|\tau)$ is a prior on the true value $y_t^i$ given the class $\tau$. Two things are worth noting: first, because of the uncertainties in the training data, the classification of just a single scalar data point requires an $n$-dimensional integral over the $n$ instances in each class $\tau$\footnote{Strictly speaking we should write $n_{\tau}$ since the number of samples in each class will be different but we suppress this to keep the notation relatively simple.}. Second, equation \ref{eq:generalcase1dp} is not just a product of terms like equation \ref{eqn:product}, even if the instances are independent. 

We next generalise to consider the case of a multi-dimensional feature vector for each instance, i.e. $\{y_{o,j}^i\}$ for the training data and $\{d_j\}$ for the test data instance, where as usual $i,j$ index the instance and data dimension within an instance respectively (e.g. for time-series data $j$ would index the time). In the special case of uncorrelated Gaussian distributed test and training data, and for (improper) flat priors, we can analytically compute the posterior probability, $P(\tau|\{d_j\},\{y_{o,j}^i\})$ (see Appendix \ref{sec:detailed-deriv}), yielding our main result:   
\begin{multline}
P(\tau|\{d_j\},\{y_{o,j}^i\}) \propto \frac{1}{n}\sum^n_{i=1} P(\tau_i) \prod_{j=1}^m (2\pi \sigma_{d_j} \sigma_{y_{o,j}^i})^{-1} \left[ \dfrac{\pi}{\frac{1}{2}(\Gamma_d + \Gamma_i)} \right]^{1/2} \\ 
\times\ \exp\left(-\frac{1}{2}\left(\Gamma_d d_j^2 + \Gamma_i {y_{o,j}^i}^2 -\dfrac{(\Gamma_d d + \Gamma_i y_{o,j}^i)}{\Gamma_d + \Gamma_i}\right)\right)
\label{eq:maineq}
\end{multline}
where $\Gamma_d \equiv \sigma_{d,j}^{-2}$ and $\Gamma_i \equiv \sigma_{y_{o,j}^i}^{-2}$ are the precision of the datapoints, $n$ is the total number of training instances and $m$ is the number of datapoints per instance. We use equation \ref{eq:maineq} in all our experiments in section \ref{sec:results} to evaluate BADAC, though it is of course not the most general form.  

To normalise equation \ref{eq:maineq} we need to compute the Bayesian evidence for all the observed data, $P(\{d_j\},\{y_{o,j}^i\})$. For $m$ different classes in total, the evidence is then:
\begin{equation}
P(\{d_j\},\{y_{o,j}^i\}) = \sum_{k}^m P(\{d_j\},\{y_{o,j}^i\}|\tau_k) P(\tau_k)
\end{equation}
This is a summation of the posterior probability terms returned for each class type which requires the prior probability, $P(\tau_k)$, for each class, to be supplied by the user as usual. Classification then occurs by choosing the class with the highest normalised posterior probability. 

\begin{figure}[h]
\includegraphics[width=.8\columnwidth]{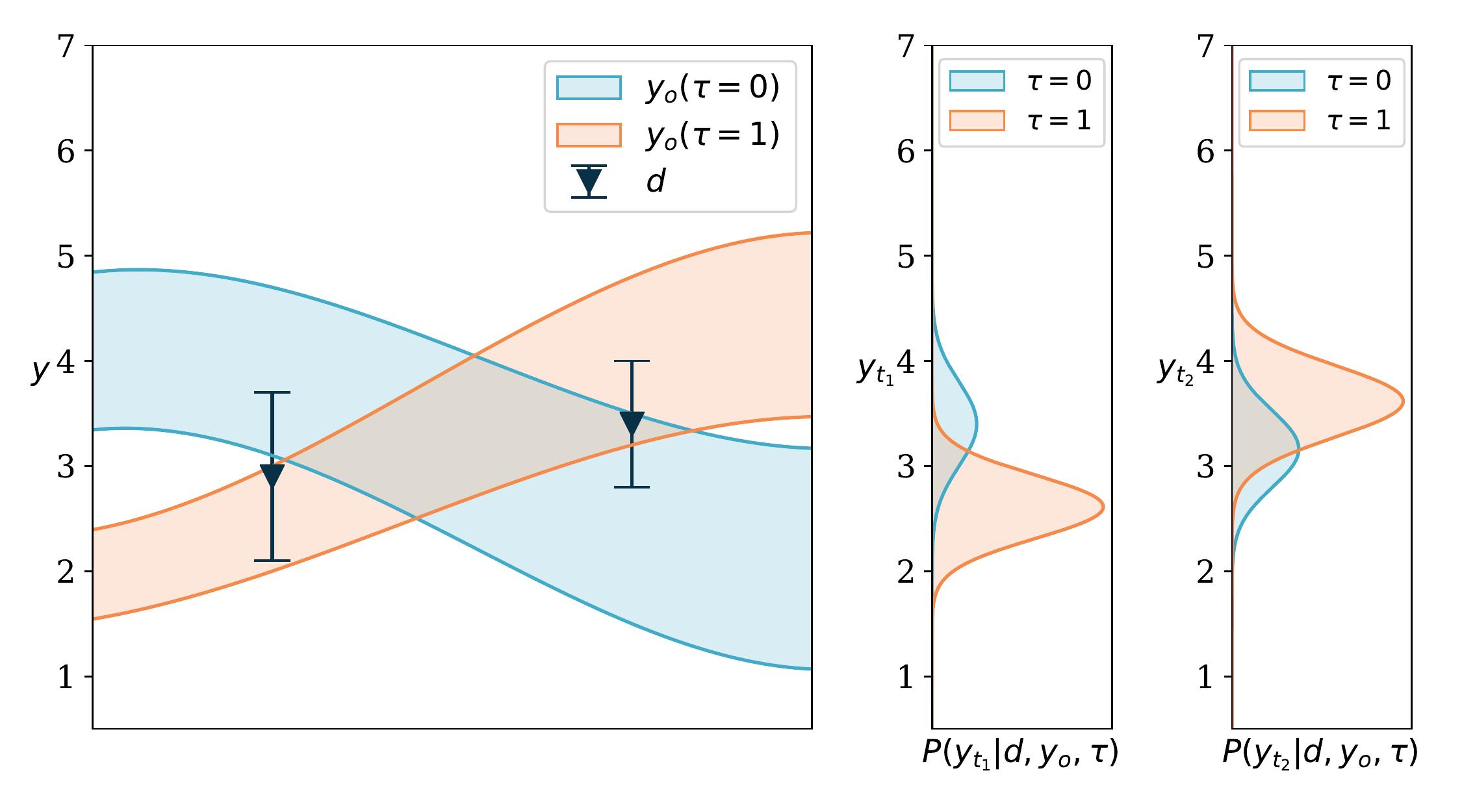}
\centering
\caption{Schematic representation of BADAC as a classifier. {\em Left}: a single test example consisting of just two data points (black triangles with error bars). The training data comes from two classes shown schematically as the blue ($\tau = 0$) and orange ($\tau = 1$) 1-$\sigma$ error envelopes. Which of these two classes does the test data come from?  {\em Middle and Right}: panels showing the unnormalised posterior probability for the true value, $y_t^i$, for the first (middle panel) and second (right panel) data point, marginalised over the true value of the other point and conditioned on belonging to either class (class 0 - blue or class 1 - orange). The relative area of the corresponding Gaussians in the middle and right panels gives the probability for the data to belong to either class. As can be seen, the data is more likely to come from class 1 (the orange class), in this case with a probability of $73\%$.
}
\label{fig:schem}
\end{figure}

We illustrate BADAC for Bayesian classification in figure \ref{fig:schem}. Here we show a toy example with training data synthesised into two smooth Gaussian templates (whose 1$\sigma$ regions are shown by the filled blue and orange areas) corresponding to two classes, $\tau = 0,1$. We have a single test example which we wish to classify, consisting of two data points (the black triangles with error bars). The middle and right panels show the unnormalised posteriors for the true values of the points, i.e. $P(y_t^i | d, \tau = 0,1)$, marginalised over the true value of the other point. For the first data point the posterior conditioned on the blue class is pulled upwards and that conditioned on the orange class downwards. However, because the blue error envelope at the first data point is broader than the orange one, the first point is less likely to come from the blue class, a fact reflected in the smaller area under the blue Gaussian in the first of the right narrow panels. The same happens for the 2nd point. The Bayesian evidence therefore predicts that the test example is more likely to come from the $\tau = 1$ (orange) class, with probability $65\%$. 

\subsection{Anomaly Detection}
Since we want a simultaneous anomaly detection and classification algorithm we assign the $m$-th class to the unknown anomalies with the known classes assigned to the remaining $m-1$ $\tau$ labels. A key issue is the choice of the likelihood for the anomaly class: $P(\{d_j\},\{y_{o,j}^i\}|\tau = m)$. If we have seen examples of the anomaly before, this can be estimated from the data. If the anomaly has never been seen before however, this reduces essentially to a prior about our expectations regarding the anomaly. Broad, normalised Gaussian or top-hat functions are simple choices allowing for a wide range of behaviours for the anomalies. The shadow of the No Free Lunch theorem appears here: if our prior about the anomaly's behaviour is completely wrong, we may miss the anomaly entirely.    

Instead of specifying a generic prior for the anomaly likelihood, an alternative approach is to assume only the known $m-1$ classes, but then to rank instances in the test data from smallest to largest values of 
\begin{equation}
\sum_{k}^{m-1} P(\{d_j\},\{y_{o,j}^i\}|\tau_k) P(\tau_k)\,,
\end{equation}
i.e. to rank the instances from least to most likely to belong to any of the $m-1$ known classes. This is useful if one has some insight into the expected fraction of anomalous events, since one can then use it to determine a cutoff in this ranked list.  

%one can instead assume a percentage contamination of the dataset, then we can determine the value of $P_{\rm anomaly}$ (the height of the top-hat).

%If for example we had $2$ known classes, then $m=3$. In this case we would get 3 probability terms for each object. Before these terms are normalised by the Bayesian evidence, they would be: 
%\begin{align}
%P_0 &= P(d,y_o^1,...,y_o^n|\tau=0) P(\tau=0)\\
%P_1 &= P(d,y_o^1,...,y_o^n|\tau=1) P(\tau=1)\\
%P_{\rm anomaly} &= P(d,y_o^1,...,y_o^n|\tau={\rm anomaly}) P(\tau={\rm anomaly})
%\end{align}

%$P_0$ and $P_1$ are calculated using the formalism shown in section \ref{sec:BHM}. We use a broad top-hat likelihood for the calculation of $P_{\rm anomaly}$. If a broad top-hat with a naively chosen width is not suitable for use as the anomaly likelihood, one can instead assume a percentage contamination of the dataset, then we can determine the value of $P_{\rm anomaly}$ (the height of the top-hat). This is done by ranking the dataset by $P_0 + P_1$, and setting $P_{\rm anomaly}$ to $P_0 + P_1$ from the $n$-th lowest summed probability object, where $n$ is the objects in the dataset that corresponds to a percentage contamination. The width of the top-hat likelihood is then determined by enforcing the condition that that the integral over the likelihood equals $1$.

\subsection{Comparison with kNN and KDE} 
It is interesting to compare BADAC to the k-nearest neighbours (kNN) algorithm \citep{altman1992}, with the best value of k selected on a validation set. Since each object can be considered as a single point in a high dimensional feature space,  in a sense BADAC generalises kNN to allow for error bars on the test and training points, and uses all the training data from each class. BADAC is also similar to the Kernel Density Estimation (KDE) approach where classification is performed by putting a kernel on each training data point and summing the resulting smooth functions. In our case the bandwidth of the kernel is given by the error bars on the data while it also handles errors on the test data. BADAC can then be thought of as a natural extension of both kNN and KDE to a more rigorous statistical setting.

\subsection{Non-Gaussian data}\label{sec:nongauss}

Often, the standard deviation is used as a proxy for the error distribution on an observation, even when the distribution is non-Gaussian. In section \ref{sec:results} we test how the algorithm developed in section \ref{sec:BHM} performs while assuming a Gaussian error distribution, even when the error distribution is non-Gaussian. However, if the error distribution is known, the forms of $P(d|y_t^i,\tau)$ and $P(y_o^i|y_t^i,\tau)$ in equation \ref{eq:generalcase1dp}, can be replaced with the known non-Gaussian distribution. These could be the binomial distribution in the case of count data, or the Poisson distribution in the case of certain time series. Any appropriate distribution that can be modelled can be used in this formalism. In the case of the binomial distribution, one would do a summation rather than an integration over the latent variables. For any distribution that doesn't yield an analytically integrable form for $P(\tau|d,y_o^1,...,y_o^n)$, one can do the marginalisation numerically, though at increased computational cost.

\subsection{Online Learning of new classes}\label{sec:online}

Once we have confirmed that an anomaly represents a new class (i.e., if the Bayesian evidence for the anomaly class is higher than that of any of the existing classes) it can automatically be added to the training data as a new class (with a single example) to be compared with. This provides an online learning version of the BADAC algorithm. Any future data belonging to the new anomaly class will be automatically assigned the new anomaly class label.  

This process in no way limits us to a single new anomaly class. The BADAC formalism allows for the automatic addition of new classes as demanded by the data. If a new kind of anomaly is different from any previously identified anomalies it will automatically be assigned to a new class. 

\subsection{BADAC with Parametric Models} \label{parametric_BADAC}

It is often the case that the training data can be efficiently modeled by some general parametrisation (with class-dependent parameters $\theta_\tau$), either because we have a physical model of the classes or have performed a general expansion on each class, e.g. through a suitable basis set such as \mlfeb{wavelets \citep{mallat:2009} or principal component analysis \citep{hotelling1933}}. As a result we assume here that we can parametrise our data $d$ as $y(\theta_\tau,x)$. Then we can express the classification probability as:
\begin{align}
P(\tau|d) & \propto \int_{\theta_\tau} d\theta_\tau P(d,\theta_\tau|\tau) P(\tau) \\
P(\tau|d) & \propto \int_{\theta_\tau} d\theta_\tau P(d|\theta_\tau,\tau) P(\theta_\tau|\tau) P(\tau)
\end{align}
Here $P(\theta_\tau|\tau)$ is the prior on the parametrised variables $\theta_\tau$, for a given class $\tau$. $P(d|\theta_\tau,\tau)$ is the likelihood for given parameters $\theta_\tau$, given data $d$. For models derived from physical phenomena, we can usually evaluate this. 

If one is using some general parametrisation, such as a wavelet transform, then the parameters $\theta_\tau$ can be thought of as similar to (and hence grouped with) $y_o$. In this case, if there were measurement uncertainties associated with each $\theta_\tau$, we could then proceed in the same way as in equation \ref{eq:generalcase1dp}, where instead of marginalising over a large range of possible $y_t$ values, we now marginalise over the more constrained $\theta_t$. This could be extended to the hierarchical context where the parameters $\theta_t$ are drawn from some parent distribution. 

This shift from the raw training data, to a parametrised model of the data, can have significant computational advantages when there is a lot of training data, as we discuss further in section \ref{sec:templates}.

\section{Rank-Weighted Score}\label{sec:RWS_score}

In this section we introduce a new anomaly-sensitive metric that we call the Rank-Weighted Score (RWS). In addition to being insensitive to class imbalance, this metric is sensitive to the relative \emph{ranking} of anomalous objects. In many cases there is a clear hierarchy of how interesting anomalous objects are (a new class is generally more interesting than a new subclass for example), and hence we prefer algorithms that correctly rank more anomalous objects more highly\footnote{Compare for example the Spearman's rank correlation \citep{spearman1904} which does not give higher weight to the top-ranking objects.}. This is natural because following up and investigating potential anomalies typically consumes resources (whether human or instrumental) and false positives, at any anomaly threshold, must be minimised. 

The RWS is defined by ranking the $N$ objects according to their degree of anomalousness (from high to low) as identified by an algorithm. Here $N$ is a user-supplied integer (the expected number of anomalies in the dataset). The RWS score is then computed as the weighted sum: 
\begin{equation}\label{eq:rws}
S_{\rm RWS} = \dfrac{1}{S_0} \sum_{i=1}^N w_i I_i
\end{equation}
where:
\begin{equation}
w_i = (N + 1 - i)
\end{equation}
Note that this gives (linearly) more weight to correctly identifying anomalies at the top of the ranks (with low values of $i$) compared to lower down the list. In equation \ref{eq:rws}, $I_i$ is an indicator variable: $I_i=1$ if the $i$-th object is an outlier, and $I_i=0$ otherwise. $S_0$ is a normalisation factor:
$S_0 = \dfrac{N}{2}(N+1)$. 
This means the RWS score has a possible range of $[0,1]$, where $0$ implies that no true outliers were found in the $N$ most anomalous objects ranked by the algorithm, while an RWS  score of $1$ would mean that \er{all $N$ most anomalous objects identified by the algorithm were in fact outliers.} The value of $N$ must be chosen on a per problem basis, and be kept consistent across the various algorithms being considered to allow fair comparison. In section \ref{sec:results} we use this metric along with several other commonly used metrics to gauge algorithm performance. We discuss these metrics in Appendix \ref{sec:metrics}.

\section{Evaluation of BADAC: Gaussian Case}\label{sec:results}

To illustrate and test the performance of BADAC, we simulate a number of one-dimensional datasets and compare results with multiple metrics including the Rank-Weighted Score (RWS) introduced in section \ref{sec:RWS_score}, that is optimised for anomaly detection. 

\subsection{Simulations}\label{sec:simulations}

We simulate data from arbitrary mathematical functions. We use two mathematical functions to build two ``normal'' classes and use three other functions as anomalies. Each function has parameters which, when generating the data, are randomly drawn from a Gaussian distribution. The class functions and their corresponding parameter distributions are given in table \ref{tab:functions}.

\begin{table}[!h]
\caption{Description of the functions used to create the simulated data. 99\% of the \mlnew{test objects in the dataset are of the type ``inlier''} and 1\% are ``outliers''. Each class has the corresonding functional form with  parameters drawn randomly for each instance from Gaussian distributions with hyperparameters specified in the table.}
\label{tab:functions}
\centering
\begin{tabular}{cccc}
\hline
Class label & Type & Functional Form & Parameter distributions\\
\hline
0 & Inlier & $y = \rm{sin}(\omega x)$ & 
\begin{tabular}{@{}c@{}}
\\
 $ \omega \sim \mathcal{N}(5,2)$\\
\\
\\
\end{tabular}\\
\hline
1 & Inlier & $y = \alpha x^2 + \beta x + \gamma$ & 
\begin{tabular}{@{}c@{}}
\\
$\alpha \sim  \mathcal{N}(0.5, 0.2)$\\
$\beta \sim  \mathcal{N}(0.5, 0.2)$\\
$\gamma \sim  \mathcal{N}(0, 0.2)$\\
\\
\end{tabular}\\
\hline
2 & Outlier & $y =  h$ if  $x \leq x_0$, else $y=0$ &
\begin{tabular}{@{}c@{}}
\\
$h \sim \mathcal{N}(1, 0.3)$\\
$x_0 \sim \mathcal{N}(0.5, 0.2)$\\
\\
\end{tabular}\\
\hline
3 & Outlier & $y = A \, \exp \left( -\bigg(\frac{x-\mu}{w} \bigg)^2 \right)$ &
\begin{tabular}{@{}c@{}}
\\
 $A  \sim  \mathcal{N}(0.5, 0.2)$\\
 $\mu  \sim  \mathcal{N}(0.1, 0.05)$\\
 $w  \sim  \mathcal{N}(1, 0.5)$\\
 \\
\end{tabular}\\
\hline
4 & Outlier & 
\begin{tabular}{@{}c@{}}
$y = 0.2\bigg(\sin(\omega_1 x) + \sin(\omega_2 x) + $\\
$\sin(\omega_3 x) + \sin(\omega_4 x)  + \sin(\omega_5 x) \bigg)$ 
\end{tabular}&
\begin{tabular}{@{}c@{}}
\\
 $\omega_i  \sim  \mathcal{N}(30, 20)$\\
\\
\end{tabular}\\
\hline
\end{tabular}
\end{table}

For each experiment, we generate $15000$ curves of roughly equal number of objects from class 0 and 1 as training data. 
In the test data, we add 1\% outliers from classes 2, 3 and 4. Figure \ref{fig:data} illustrates some randomly drawn objects from the training and test sets (equal numbers from each class).

~\\
\begin{minipage}{0.99\columnwidth}
\centering
\includegraphics[width=0.7\columnwidth]{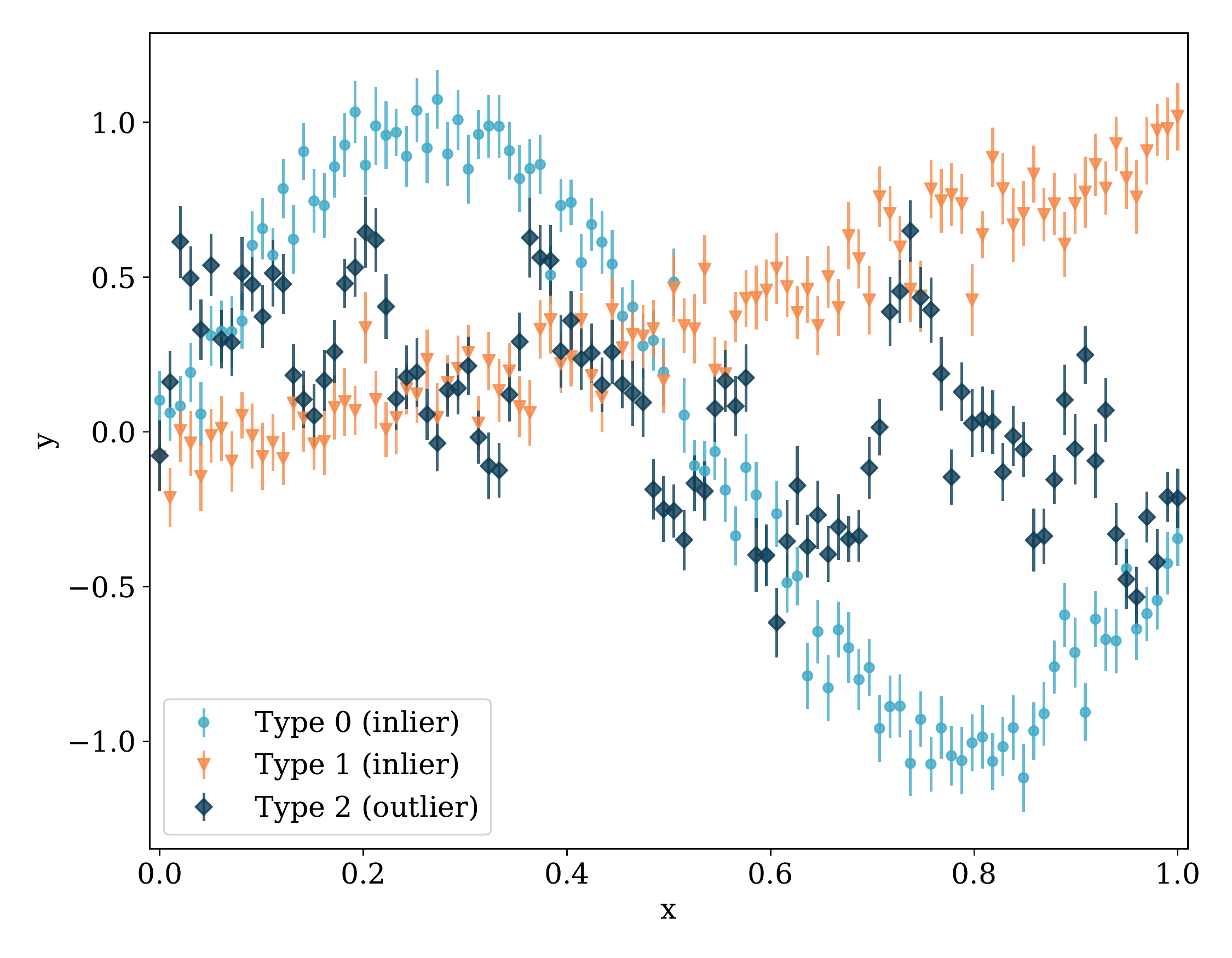}
\captionof{figure}{Illustrations of example objects from the \mlfeb{simulated} data. The plotted error bars correspond to $1\sigma$ error of Gaussian noise. The functional form and distribution of hyperparameters used to generate these examples is shown in in table \ref{tab:functions}. The points are coloured by true type, where light blue circles correspond to a type 0 object, orange triangles to type 1 and dark indigo diamonds is an outlier. Only type 0 and type 1 curves are used during the training phase.}
\label{fig:data}
\end{minipage}

\subsubsection{Experiment 1: Gaussian errors} \label{sec:experiment-gauss}

We use the framework of section \ref{sec:simulations} to create a variety of experiments to test our anomaly detection and classification algorithm. Here we simulate the data as described in section \ref{sec:simulations} with uncorrelated Gaussian errors on all data points. The standard deviation of the underlying noise distribution depends on the class, and is given by: $\sigma_0 = \sigma_2 = \sigma_3 = \sigma_4 = 0.3$ and $\sigma_1 = 0.5$. This experiment is the ideal case in which the noise distribution used for generating the simulated data is the same as that in the mathematical formulation of equation \ref{eq:maineq}.  

\subsubsection{Experiment 2: Compact Anomalies} \label{sec:experiment-CA}
In this experiment, we test BADAC's ability to detect curves with a compact anomaly embedded somewhere in them. We use one of the base inlier classes described in experiment 1, the sine curve (class 0), and place on top of it a narrow Gaussian. We randomly draw the parameters of the sine curve from the distribution described in table \ref{tab:functions} and draw the parameters of the compact anomalies as described in table \ref{tab:anomalies}. An example of a compact anomaly is shown in figure \ref{fig:anomalies}.

\begin{table}[!h]
\caption{Description of the functions used to create the compact anomaly simulated data. 99\% of the test objects in the dataset are of the type ``inlier" which are the same as class 0 and 1 in table \ref{tab:functions}. The remaining 1\% are drawn from one of two compact anomaly classes. These are narrow Gaussians added to a randomly generated function of class 0. The parameters of the Gaussian are drawn randomly for each object from a distribution with hyperparameters as specified in the table.}
\label{tab:anomalies}
\centering
\begin{tabular}{cccc}
\hline
Class label & Type & Functional Form & Parameter distributions\\
\hline
0 & Inlier & $y = \rm{sin}(\omega x)$ & 
\begin{tabular}{@{}c@{}}
\\
 $ \omega \sim \mathcal{N}(5,2)$\\
\\
\\
\end{tabular}\\
\hline
1 & Inlier & $y = \alpha x^2 + \beta x + \gamma$ & 
\begin{tabular}{@{}c@{}}
$\alpha \sim  \mathcal{N}(0.5, 0.2)$\\
$\beta \sim  \mathcal{N}(0.5, 0.2)$\\
$\gamma \sim  \mathcal{N}(0, 0.2)$\\
\\
\end{tabular}\\
\hline
2 & Outlier & $y = \rm{sin}(\omega x) + A \, \exp \left( -\bigg(\frac{x-\mu}{w} \bigg)^2 \right)$ &
\begin{tabular}{@{}c@{}}
  $ \omega \sim \mathcal{N}(5,2)$\\
 $A  \sim  \mathcal{N}(1.5, 0.5)$\\
 $\mu  \sim  \mathcal{U}(0, 1)$\\
 $w  \sim  \mathcal{N}(0.03, 0.01)$
\end{tabular}\\
\hline
3 & Outlier & $y = \rm{sin}(\omega x) + A \, \exp \left( -\bigg(\frac{x-\mu}{w} \bigg)^2 \right)$ &
\begin{tabular}{@{}c@{}}
 $ \omega \sim \mathcal{N}(5,2)$\\
 $A  \sim  \mathcal{N}(-1.5, 0.5)$\\
 $\mu  \sim  \mathcal{U}(0, 1)$\\
 $w  \sim  \mathcal{N}(0.03, 0.01)$
\end{tabular}\\
\hline
\end{tabular}
\end{table}
~\\
\begin{minipage}{0.99\columnwidth}
\centering
\includegraphics[width=0.75\columnwidth]{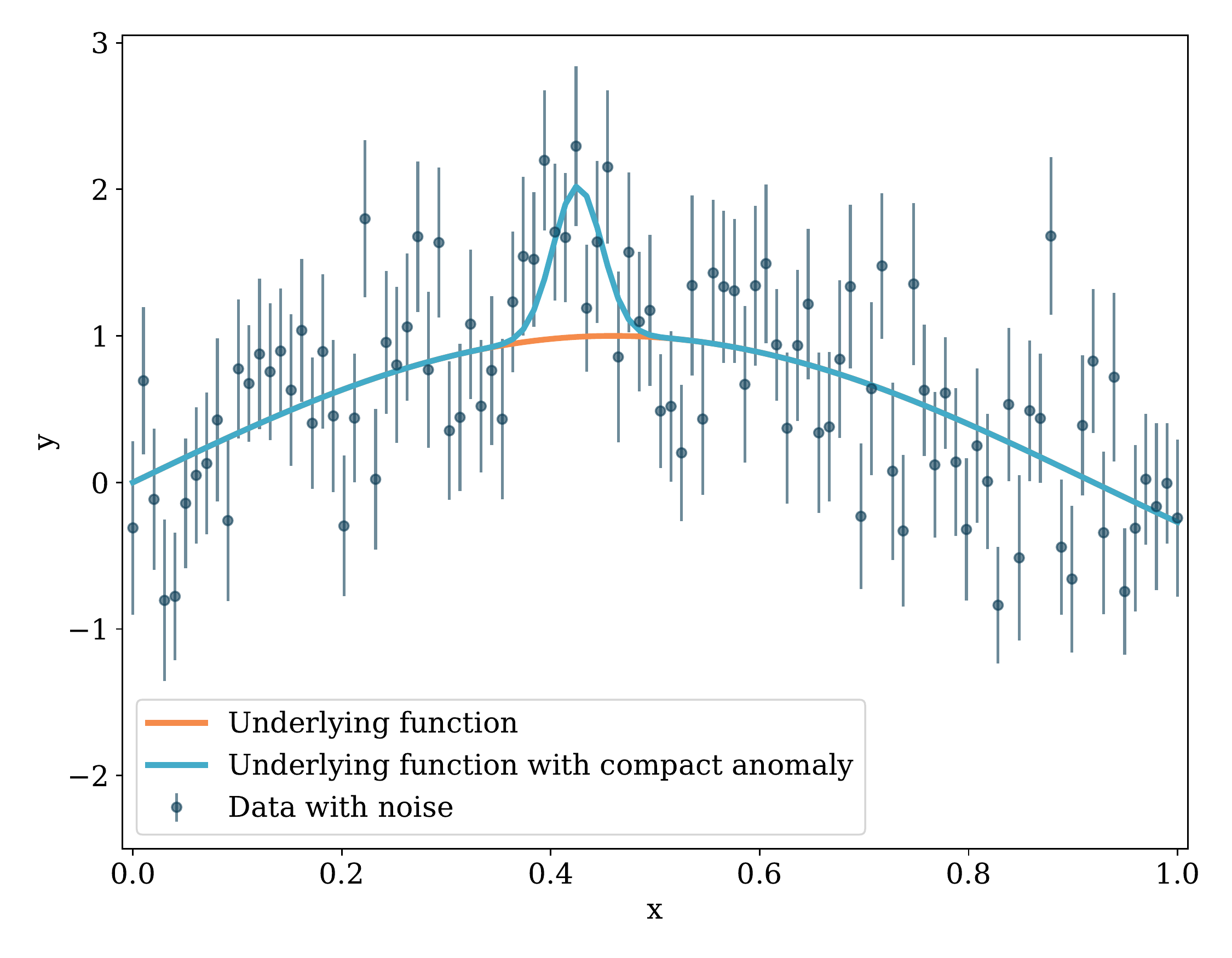}
\captionof{figure}{Example of the compact anomaly simulations. The underlying function from which the data were generated is shown as an orange solid line. The underlying function with the compact anomaly superposed is shown as the light blue solid line. The final data with noise are shown by the dark indigo scatter where the errorbars represent the $1\sigma$ Gaussian measurement error.}
\label{fig:anomalies}
\end{minipage}

\subsection{Comparison of algorithm performance}\label{sec:algo-perform}

We assess the performance of our algorithm on the simulated data discussed in section \ref{sec:simulations}. We then compare our algorithm to a series of benchmark algorithms, namely IsolationForest \citep{liu2008} and Local Outlier Factor (LOF) \citep{breunig2000} for anomaly detection, and random forests \citep{breiman2001} for classification.  

We use \texttt{sklearn} \citep{sklearn2011} implementations for all of the benchmark algorithms we compare against BADAC. For anomaly detection, all algorithms receive only the input training data, and the percentage of outliers of $1\%$. For classification with random forests, we set the input parameter \texttt{n\_estimators=1000}. \er{There are unsupervised implementations of IsolationForest and LOF, but here we consider the supervised methods only.}

\begin{table}[!h]
\caption{Anomaly detection challenge results for the Gaussian noise simulations and Gaussian noise compact anomalies (Comp. Anom.) for the three metrics (MCC, AUC and RWS) discussed in section \ref{sec:metrics}. The best performer is shown in bold. Note the particularly poor performance of IsolationForest in the MCC and RWS metrics. BADAC significantly outperforms the other algorithms in the Gaussian case.}
\label{tab:result-summary}
\centering
\begin{tabular}{l c c c c c c c c c}
\hline
 & \multicolumn{3}{c}{BADAC} & \multicolumn{3}{c}{IsolationForest} & \multicolumn{3}{c}{LOF} \\
\hline
 & MCC & AUC & RWS  & MCC & AUC & RWS & MCC & AUC & RWS \\
\hline
Gaussian & {\bf 0.95} & {\bf 0.99} & {\bf 0.99} & 0.00 & 0.89 & 0.02 & 0.83 & 0.97 & 0.96 \\
Comp. Anom. & 0.41 & {\bf 0.91} & 0.59 & 0.11 & 0.80 & 0.14 & {\bf 0.44} & 0.90 & {\bf 0.63} \\
\hline
\end{tabular}
\end{table}

\begin{table}[!h]
\caption{Comparison of BADAC's {\em classification} performance to that of random forests using average accuracy across \er{both inlier} classes.}
\label{tab:result-summary-classification}
\centering
\begin{tabular}{p{7.5cm} c c}
\hline
 & BADAC & Random Forests \\
\hline
Gaussian Noise & {\bf 99.02} & 98.66 \\
Compact Anomalies & {\bf 95.51} & 95.18 \\
\hline
\end{tabular}
\end{table}

\subsubsection{Gaussian noise}\label{sec:algo-perform-gauss}

Here we illustrate the performance of BADAC, as well as the benchmark algorithms, on the data discussed in section \ref{sec:experiment-gauss} with Gaussian measurement error. We use the formalism shown in section \ref{sec:BHM} to provide two probabilities, $P_0$ and $P_1$, which are the un-normalised probabilities of belonging to class 0 and class 1 respectively. These probabilities are plotted in figure \ref{fig:scatter-gauss}.
~\\
\begin{minipage}{0.99\columnwidth}
%\vspace{10pt}
\centering
\includegraphics[width=.7\columnwidth]{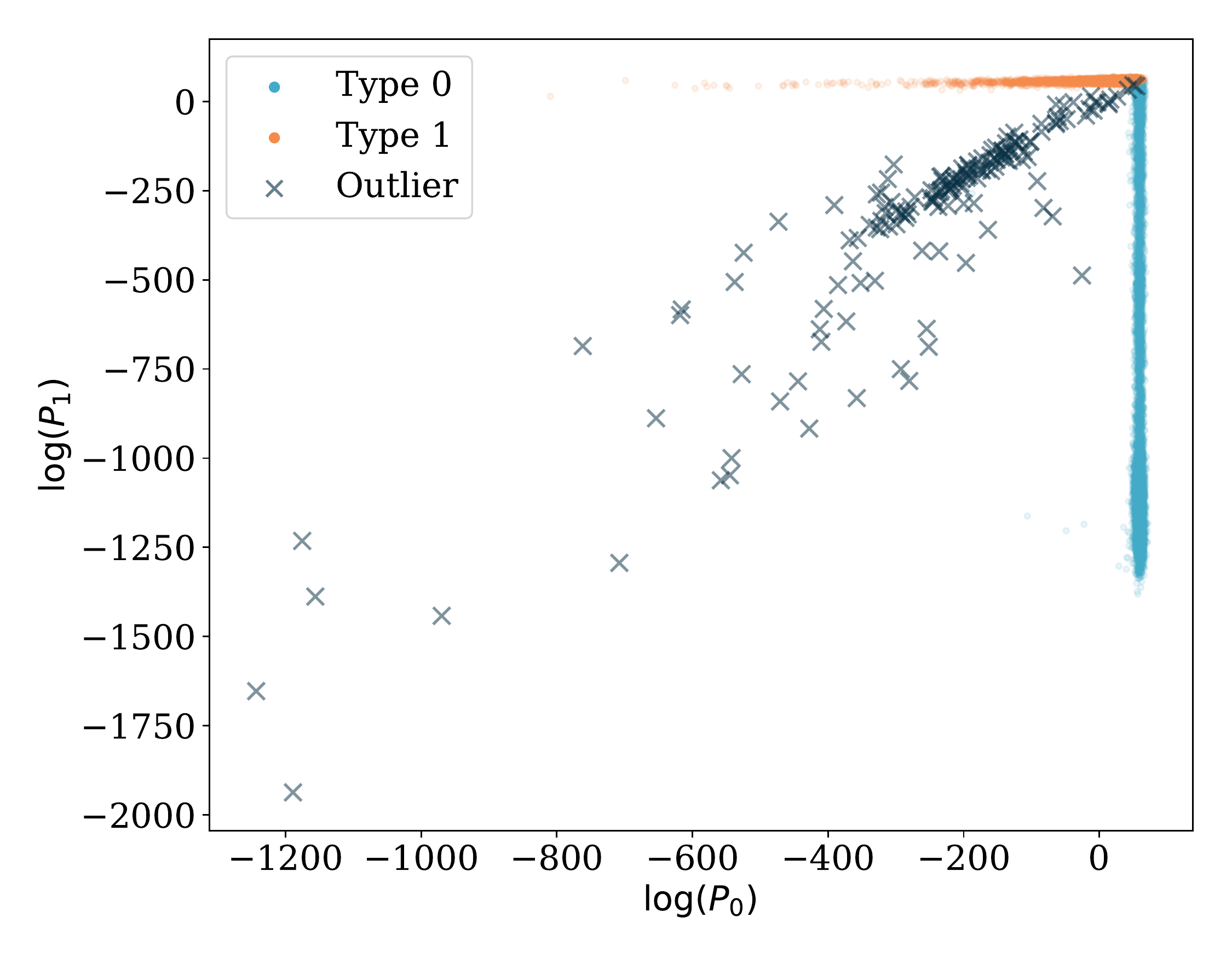}
\captionof{figure}{Scatter plot showing the computed log-probabilities for the test data discussed in section \ref{sec:experiment-gauss}. Each point corresponds to a test object, which is shown in the log($P_0$)-log($P_1$) space. Points that appear high on the y-axis have a high likelihood of being type 1. Points that appear higher (to the right) on the $x$-axis have a high likelihood of being type 0. The points are coloured by true type, where light blue corresponds to type 0, orange is type 1 \er{and the dark crosses are all outliers.}}
\label{fig:scatter-gauss}
\end{minipage}
%\vspace{10pt}

\er{Plotting the unnormalised probabilities is useful for visualising the decision boundary that separates both the known classes and anomalies. It also does not require us to make any assumptions about the nature of the anomalies we expect to see. \mlfeb{However, to make use of these probabilities in an analysis pipeline, they must be normalised.} In order to normalise the probabilities, we compute the Bayesian evidence.} The evidence in the case where one is interested in classification only would be $P_0+P_1$. In the case where anomaly detection is of interest as well, the evidence is $P_0+P_1+P_{anomaly}$, where \er{in this case,} we choose to evaluate $P_{anomaly}$ using a top-hat likelihood equal to $1/(b-a)$ over the range of $[a,b]$, and equal to $0$ otherwise. Here we choose $a$ and $b$ to cover twice the observed range of the input data.

If we bin the normalised probabilities for a single class, we can measure whether or not they are calibrated. It is a well known problem that many machine learning algorithms give uncalibrated probabilities that do not correspond to the true probability of an object \er{belonging to} a certain class. The reliability of probabilities can be investigated by plotting a probability calibration curve: the output probabilities from the algorithm \er{for a selected class only} are binned and compared with the actual fraction of objects in that bin belonging to the class. We show this result for classification only for type 1 objects in figure \ref{fig:true_prob-gauss}, and compare the results of BADAC with those of random forests. 
~\\
\begin{minipage}{0.99\columnwidth}
\centering
\includegraphics[width=.7\columnwidth]{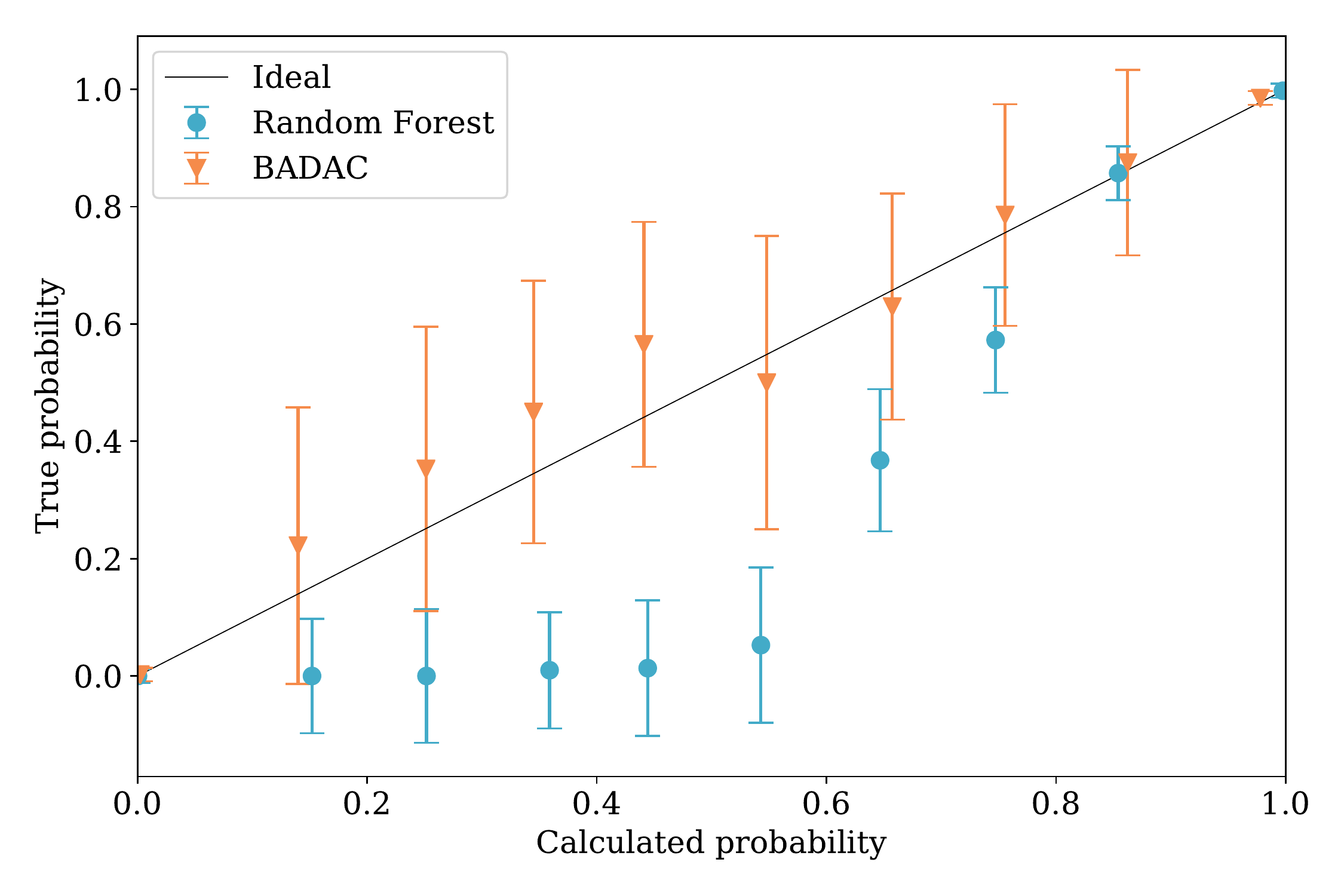}
\captionof{figure}{Probability calibration curve for the Gaussian case for BADAC and random forests (for classification only). Perfectly calibrated probabilities would lie on the line $y=x$. Here we consider the probability of an algorithm classifying an object as type 1. All objects within a particular probability range are binned, and the fraction of correct \er{positive} predictions plotted. The errorbars show the Poisson uncertainties given by the number of objects in each bin, \er{and the $x$-coordinate for each bin is given by the mean calculated probability for that bin}. Random forest gives poorly calibrated probabilities while BADAC \mlfeb{automatically} returns well-calibrated probabilities. This is to be expected since the model we use accounts for the Gaussian noise in the data and follows a principled Bayesian approach.}
\label{fig:true_prob-gauss}
\end{minipage}
~\\
\begin{minipage}{0.99\columnwidth}
\vspace{10pt}
\centering
\includegraphics[width=.7\columnwidth]{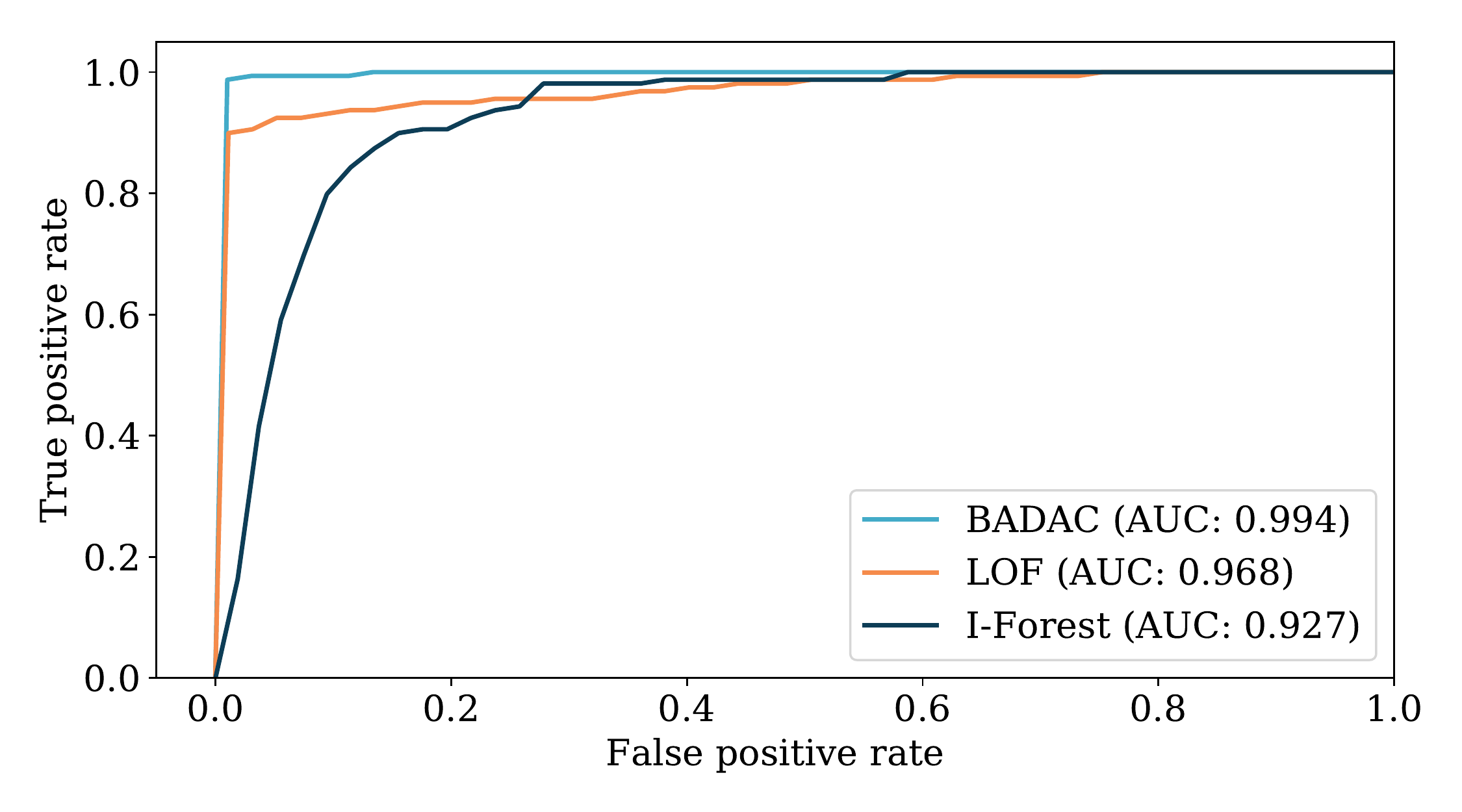}
\captionof{figure}{ROC curves for BADAC, LOF and IsolationForest on the dataset with uncorrelated Gaussian error, for anomaly detection. BADAC performs best under the AUC metric shown in each legend. The best classification algorithms have a ROC curve that reaches close to the top left hand corner, \mlfeb{with perfect performance} corresponding to an AUC of one.}
\label{fig:roc-gauss}
\end{minipage}

We show the ROC curves (see \autoref{sec:metrics} for a description of ROC curves) for BADAC as well as LOF and IsolationForest in figure \ref{fig:roc-gauss} in order to compare algorithm performance in anomaly detection. A summary of algorithm performance from all algorithms on all the datasets we consider in both anomaly detection and classification is shown in tables \ref{tab:result-summary} and \ref{tab:result-summary-classification}.

\subsubsection{Compact anomaly Performance}
In this section we illustrate the performance of BADAC as well as the benchmark algorithms we consider on the compact anomaly data discussed in section \ref{sec:experiment-CA}. It should be noted that the compact anomaly data is generated with Gaussian noise, which is the type of noise we assume in this implementation of our formalism\er{, and is also the same kind of noise as the data described in section \ref{sec:experiment-gauss}. This means we would expect the algorithms to have similar performance in {\it classification} only in this section as in section \ref{sec:algo-perform-gauss}. For this reason, we don't discuss classification performance of any of the algorithms on the compact anomaly dataset. We proceed in the exact same manner as we did in section \ref{sec:algo-perform-gauss}, except here we test how robust the algorithms are to different types of anomalies (compact ones).}

The importance of an algorithm being able to detect compact anomalies is twofold. Firstly, compact anomalies are often interesting in science when one wishes to measure or detect aberrant behaviour of known sources. Secondly, an algorithm's ability to detect compact anomalies demonstrates its overall sensitivity in measuring small variations within data.

The probabilities, $P_0$ and $P_1$, generated by the formalism we discuss in section \ref{sec:BHM}, are shown in figure \ref{fig:scatter-ca}.
~\\
\begin{minipage}{0.99\columnwidth}
\centering
\includegraphics[width=.7\columnwidth]{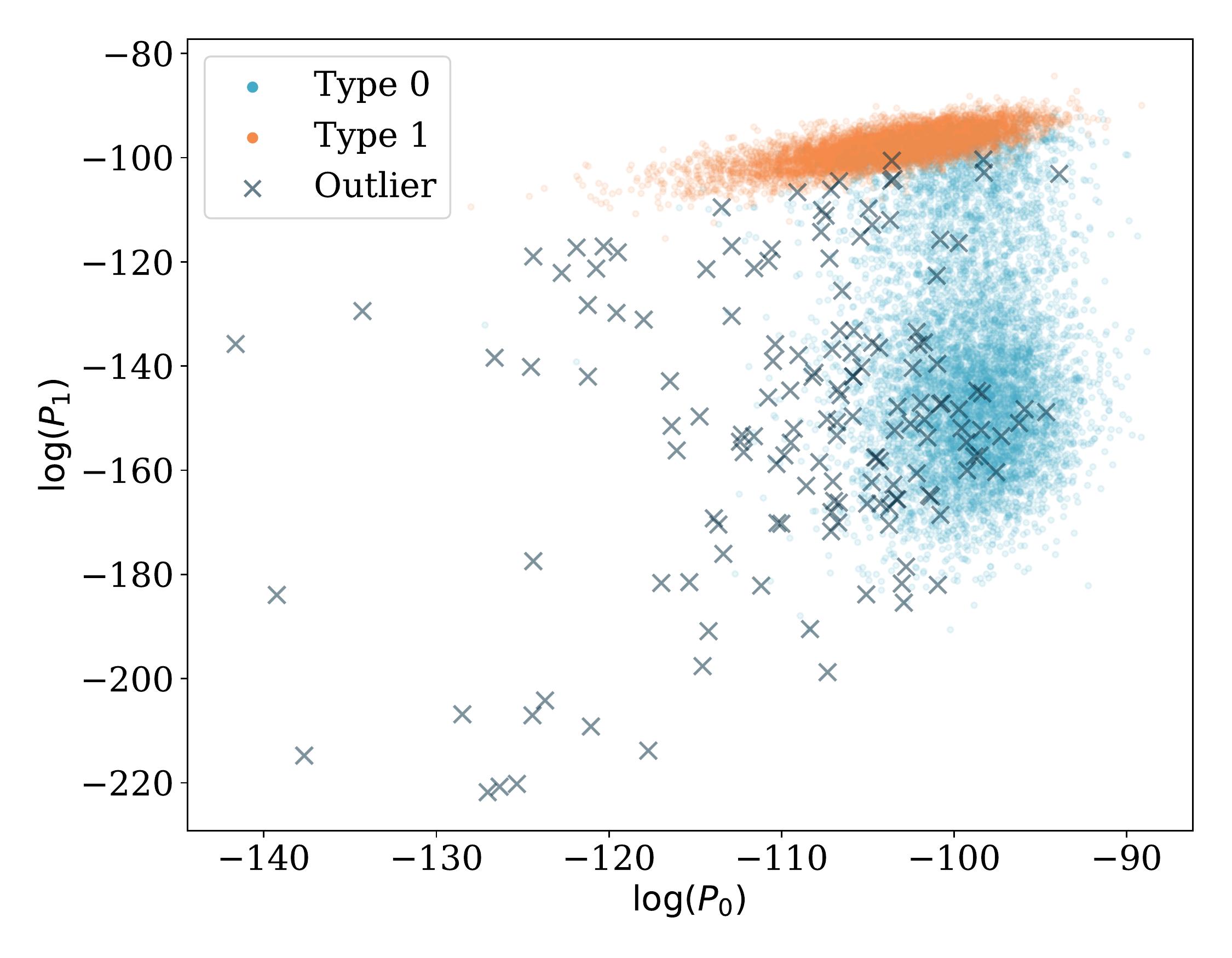}
\captionof{figure}{Scatter plot showing the computed log-probabilities for the test data discussed in section \ref{sec:experiment-CA}. Each point corresponds to a test object, which is shown in the log($P_0$)-log($P_1$) space. Points that appear high on the $y$-axis have a high likelihood of being type 1. Points that appear higher (to the right) on the $x$-axis have a high likelihood of being type 0. The points are coloured by true type, where light blue corresponds to type 0, orange is type 1 and \er{the dark crosses are outliers}.}
\label{fig:scatter-ca}
\end{minipage}

As we can see from figure \ref{fig:scatter-ca}, the outlier data has significant overlap with type 0 data. This is because we create compact anomalies on top of type 0 data only. The varying scale/amplitude to the anomaly is responsible for where the outlier data is positioned on the log($P_0$)-axis (further left indicates a more anomalous object). Outlier points with high log($P_0$) values are likely associated with compact anomalies with very low amplitudes.
~\\
\begin{minipage}{0.99\columnwidth}
\centering
\includegraphics[width=.7\columnwidth]{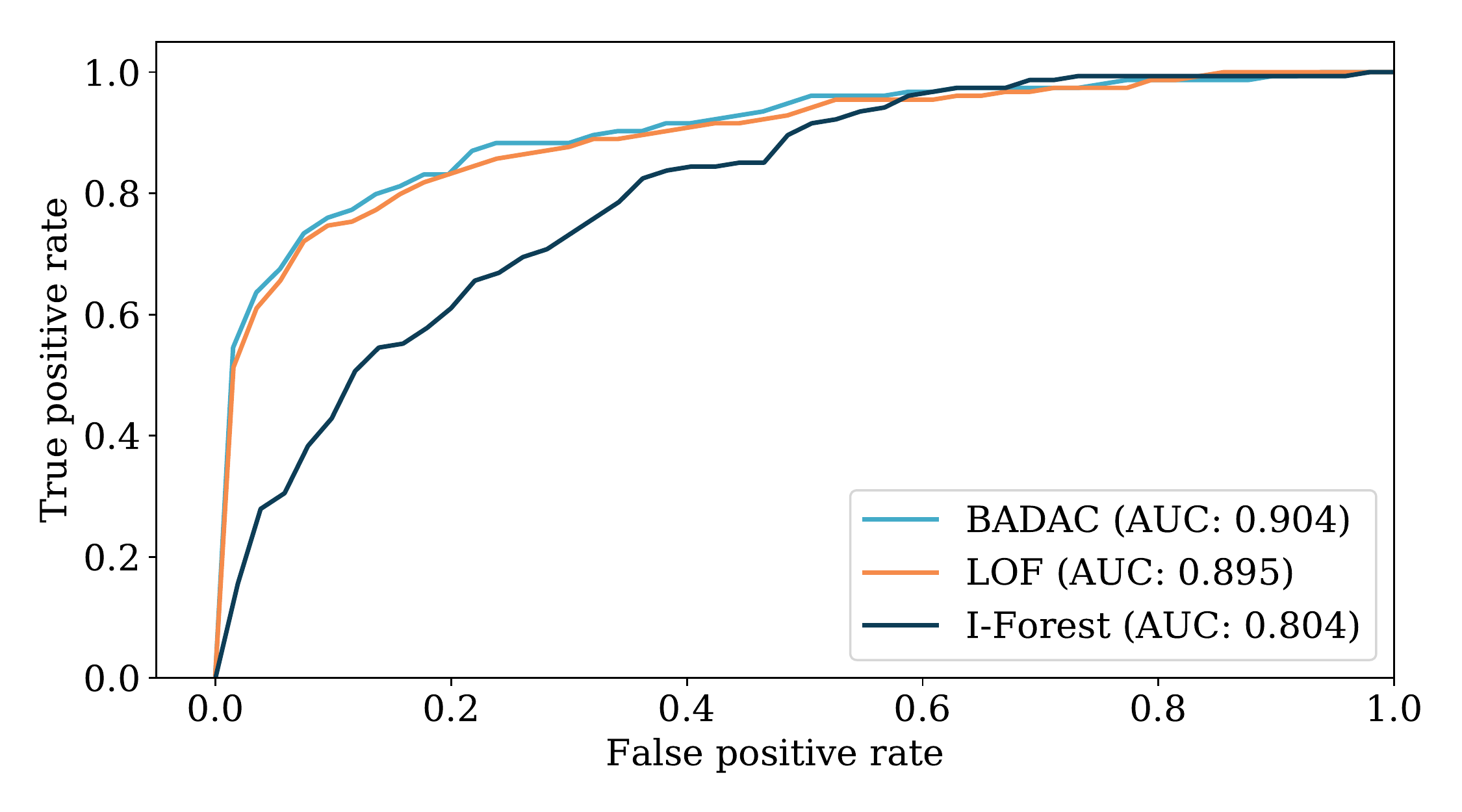}
\captionof{figure}{ROC curves for \mlfeb{anomaly detection with} BADAC, LOF and IsolationForest on the dataset with compact anomalies. BADAC performs best under the AUC metric, whose values in each case are shown in the legend.}
\label{fig:roc-ca}
\end{minipage}

We show the ROC curves for BADAC as well as LOF and IsolationForest in figure \ref{fig:roc-ca} in order to compare algorithm performance in anomaly detection. Under the AUC metric, BADAC performs the best in this case. LOF is almost as good, and actually performs better under the MCC and RWS metrics.  A summary of algorithm performance from all algorithms on all the datasets we consider in both anomaly detection and classification is shown in tables \ref{tab:result-summary} and \ref{tab:result-summary-classification}.

\subsection{Computational performance}\label{sec:computational}

It is difficult to give a ``fair" comparison of computational performance between BADAC, random forests, LOF and IsolationForest, \er{since unlike the benchmark algorithms we compare it with, our algorithm has no distinct training and testing phases}. This means that these algorithms \er{scale very differently (depending on amount of training and test data available). For example,} for a dataset with $m$ training and $n$ test examples, the computational time required for \er{random forests,} IsolationForest and LOF would increase as $f(m)+f(n)$. For our algorithm, the computational time required increases as $f(n\times m)$. In fact the computational time increases linearly as a function of $n\times m$.
%(i.e. $t_{comp} = a(n\times m)$, where $a$ is a constant). %% I feel like most readers would know what linearly means... ML
For an even comparison of computational performance, we have compared the same number of training and testing samples as were used in section \ref{sec:algo-perform} (15000 training samples and 15000 test samples). We quote the total time (training time + testing time) in table \ref{tab:time}. \mlfeb{We note, however, that there is ample room for optimisation and parallelisation in our BADAC code and the timings could be considerably improved.}

\begin{table}[!h]
\caption{Comparison of the computational performance between the three algorithms we compare in section \ref{sec:results}. All measurements were made on the dataset used in experiment 1 (Gaussian noise) with 15000 training and 15000 test curves. There are no values shown for testing and training times for BADAC, since there are no distinct training and testing phases. \er{Measurements were made on a $2.9 {\rm GHz}$ processor, where each algorithm was limited to use a single core.}}
\label{tab:time}
\centering
\begin{tabular}{p{4cm} r r r}
\hline
Algorithm & Training time (s) & Testing time (s) & Total time (s) \\
\hline
Random Forests & 96.30 & 2.94 & 99.24 \\
IsolationForest & 1.62 & 1.21 & 2.83 \\
Local Outlier Factor & 13.21 & 27.25 & 40.46 \\
BADAC & - & - & 1281.82 \\
\hline
\end{tabular}
\end{table}

As is evident in table \ref{tab:time}, BADAC has a computational cost of around an order of magnitude more than any of the competing algorithms we considered. We discuss ways of mitigating this in section \ref{sec:templates}.

\section{Breaking BADAC}

So far, i.e. in the case of Gaussian errors, BADAC has \er{outperformed random forests, LOF and IsolationForest under most metrics we consider.} This is perhaps not surprising since BADAC was designed to use the extra information available, namely that there are uncorrelated errors on the data that are Gaussian distributed. Here we try a series of more challenging tests where we use the uncorrelated Gaussian BADAC formalism, but test it on data that do not obey this model.  

\subsection{Experiment 3: Non-Gaussian errors} \label{sec:breaking-non-gauss}

Here we simulate the data exactly as in Experiment 1, however we use non-Gaussian errors instead of the Gaussian errors of Experiment 1. For 80\% of the y values (randomly selected) of any given simulated object, the noise is drawn from a Gaussian distribution with standard deviation as described in section \ref{sec:simulations}, meaning the scatter matches the error bar. However, for the remaining 20\% of the values, the noise is drawn from a Gaussian distribution of five times the width, resulting in scatter dramatically underestimated by the reported error bar.

\subsection{Experiment 4: Correlated Gaussian Noise} \label{sec:breaking-c-gauss}

To test the sensitivity of our algorithm to the uncorrelated noise assumption, we generate correlated Gaussian data. We choose to only correlate Class 0, according to a ``wedding cake" covariance matrix (based on \cite{kim2011} and \cite{knights2013}):
\begin{equation}
C_{ij} = \sigma_i \sigma_j \delta_{ij} + V_{ij},
\label{eq:wedding}
\end{equation}
where 
\begin{equation}
V_{ij} =  \displaystyle \sum \limits_{k=1}^{n_{i,j}} s_k  
\end{equation}
where $i$ and $j$ are indices of the data (in order of $x$ value) and $n_{i,j}$ is the bin to which the object belongs. To produce the step-like structure, $n_{i,j}=\lfloor 
\frac{min(i,j)}{N/5} \rfloor +1$ (where ``$\lfloor \rfloor$'' indicates the floor function, rounding down to the nearest integer). We use $s_k=0.1$ for each $k$ in this work. The result is that the data are correlated in such a way that the points at higher $x$-values are more correlated than the lower ones.
~\\
\begin{minipage}{0.99\columnwidth}
\centering
\includegraphics[width=9cm]{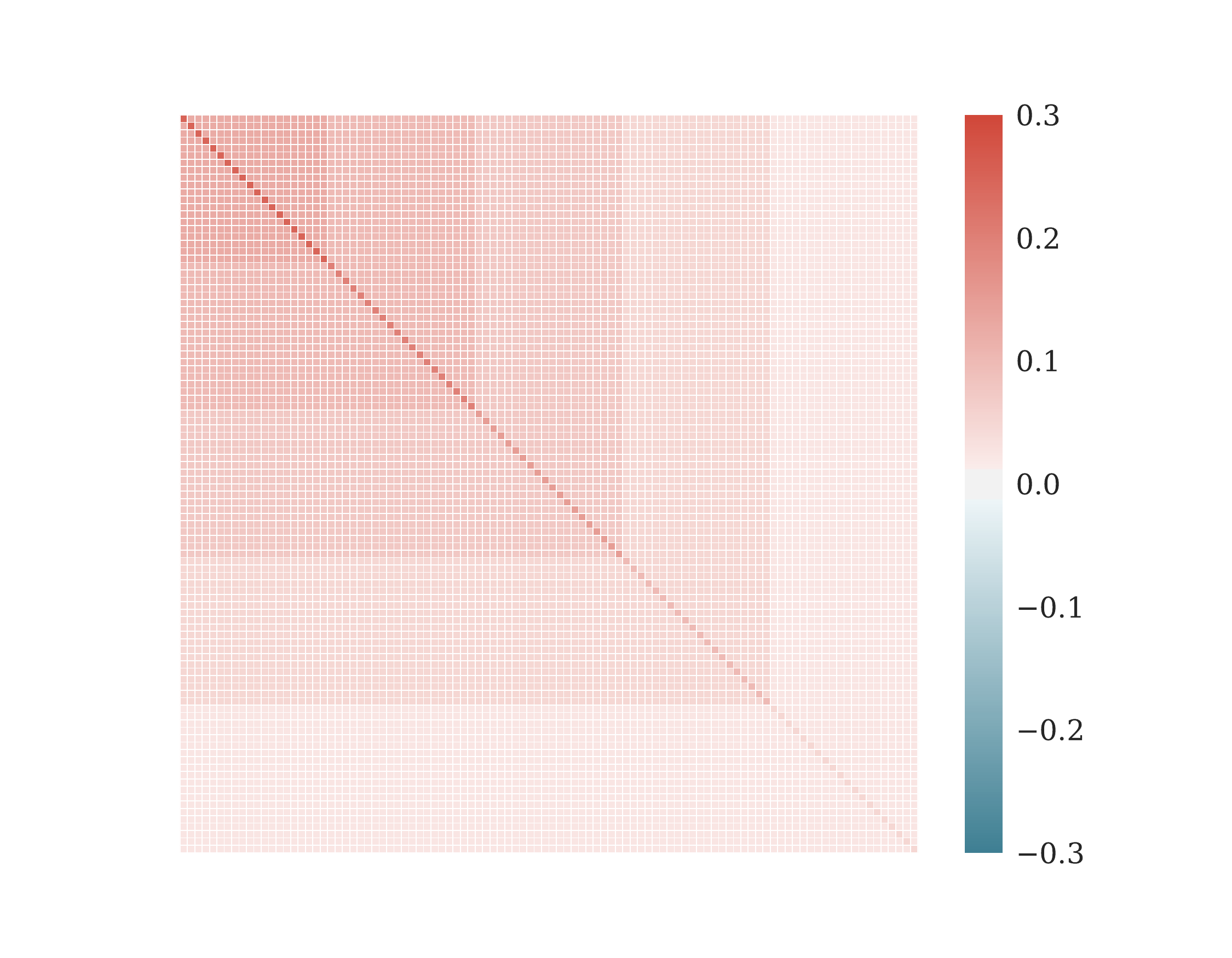}
\captionof{figure}{The covariance matrix used for correlating the class 0  data for experiment 3. This is a ``wedding cake" covariance matrix, the form of which is shown in equation \ref{eq:wedding}. The data are ordered by $x$-value starting at the top left corner (so values near the beginning of a given curve would be more highly correlated than those near the end). Class 1 and anomaly data remain uncorrelated.}
\label{fig:cov-matrix}
\vspace{10pt}
\end{minipage}
~\\
\begin{minipage}{0.99\columnwidth}
\centering
\includegraphics[width=\columnwidth]{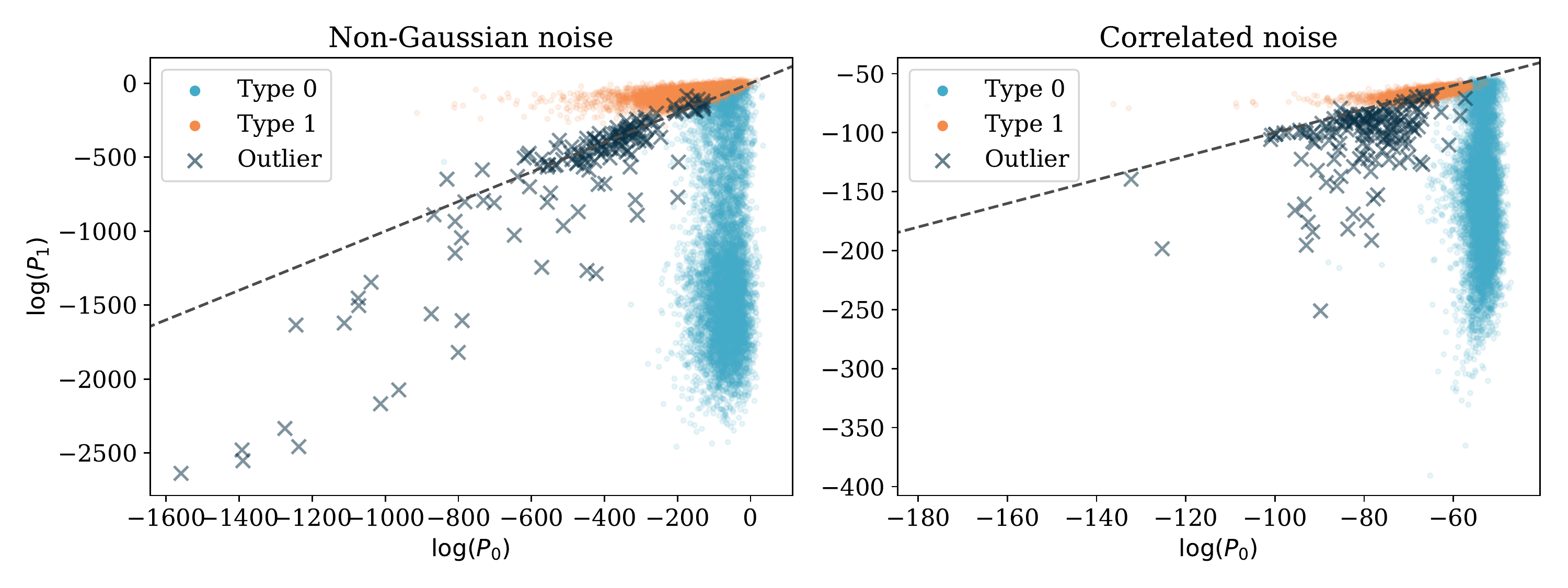}
\captionof{figure}{Probability scatter plot for the dataset with non-Gaussian noise (left panel), and the dataset with correlated Gaussian noise (right panel). Each point corresponds to a test curve, which is shown in the $\log({\rm P}0)-\log({\rm P}1)$ space. The line $y=x$ has been added to each plot to highlight the bias introduced by using the wrong model for the noise with BADAC. Here the bias is only visible in the correlated noise case since only class 0 was correlated.}
\label{fig:scatter-subplot2}
\end{minipage}

\subsection{Results for non-Gaussian noise and correlated Gaussian noise} \label{sec:results-nongauss-corr}

Here we present the results for both classification and anomaly detection for the data discussed in sections \ref{sec:breaking-non-gauss} and \ref{sec:breaking-c-gauss} with both non-Gaussian and correlated Gaussian noise. It should be noted that we still use equation \ref{eq:maineq} to determine classification/anomaly detection probabilities, despite the fact that the data does not have Gaussian uncorrelated noise as equation \ref{eq:maineq} assumes. Thus we must expect BADAC performance to decrease; the question is how much?

In order to normalise the probabilities shown in figure \ref{fig:scatter-subplot2}, we compute the evidence, $P(\{d_j\},\{y_{o,j}^i\}) = P_0+P_1+P_{anomaly}$ once again. As before, we choose to evaluate $P_{anomaly}$ using a top-hat likelihood equal to $1/(b-a)$ over the range of $[a,b]$, and equal to $0$ otherwise. In this case, naively choosing the width of the top-hat does not work, as the model used to determine $P_0$ and $P_1$ is incorrect, and hence returns low probabilities. As a result, $P_{anomaly}$ is a much higher probability than $P_0$ and $P_1$, even for inlier data, when the incorrect model for the noise is used. To get around this we determine the height of the top-hat likelihood, $1/(b-a)$, by equating it to $P_0+P_1$ computed for the object corresponding the 99th percentile. What this means is, we enforce that the algorithm labels the most anomalous 1\% of objects (as determined by the algorithm) as outliers. This is still a fair comparison with the benchmark algorithms, as both IsolationForest and LOF receive the percentage contamination of 1\% as an input parameter. We discuss how to extend this method to be suitable for modelling data with different types of noise in section \ref{sec:nongauss}.
~\\
\begin{minipage}{0.99\columnwidth}
\centering
\includegraphics[width=\columnwidth]{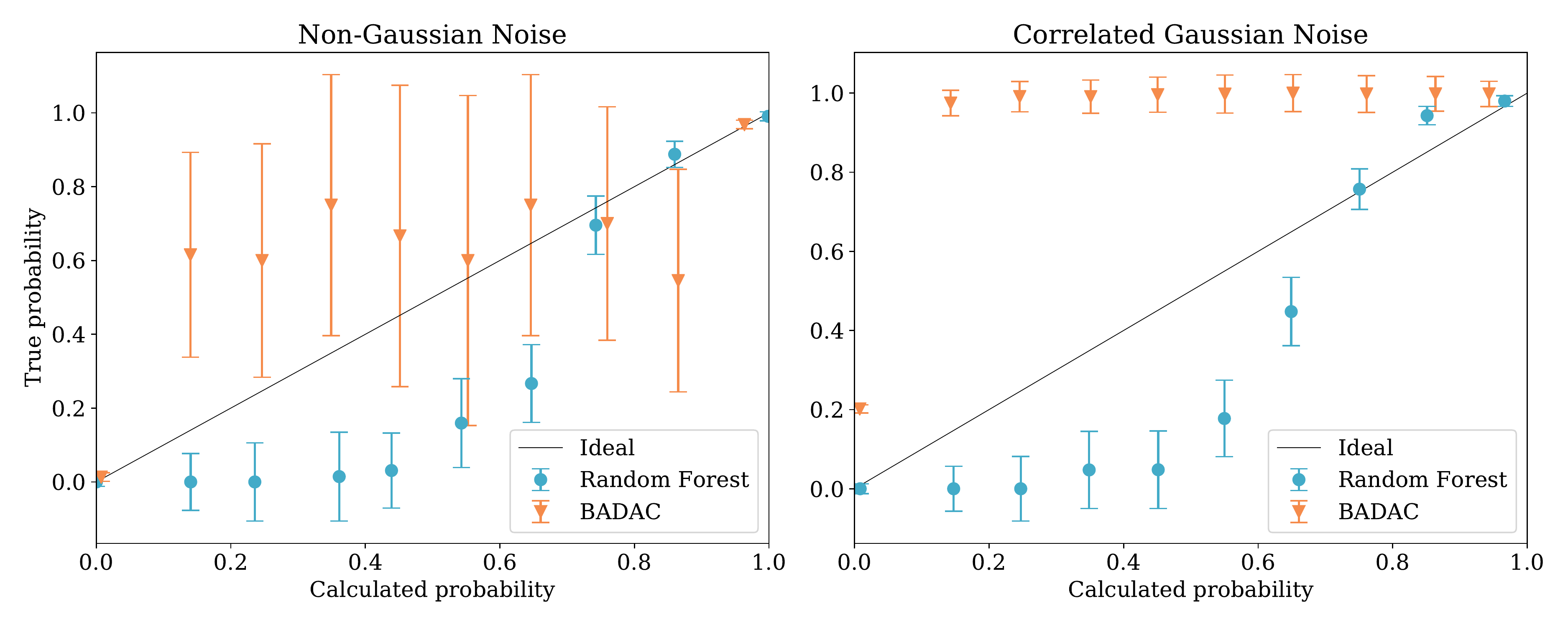}
\captionof{figure}{Probability calibration curves showing the degree to which the probabilities returned by each algorithm (in classification only) are calibrated for the non-Gaussian case we consider (left panel) and the correlated Gaussian case (right panel) respectively. Perfectly calibrated probabilities would lie on the line $y=x$. Here we consider the probability of an algorithm classifying an object as type 1. All objects within a particular probability range are binned, and the fraction of correct predictions plotted. The errorbars show the Poisson uncertainties given by the number of object in each bin. While non-Gaussian noise does not distort the probabilities dramatically, correlated noise has a strong effect due to a fundamentally incorrect noise model assumption.}
\label{fig:true-prob-subplot2}
\end{minipage}

As we can see from the scatter of classification probabilities shown in figure \ref{fig:scatter-subplot2}, there is more overlap of different object types present than in the uncorrelated Gaussian noise case we considered. Additionally, in the correlated case, there is a significant bias introduced due to the noise from only one of the classes being correlated. Since the model does not favour fitting this class, the classification probabilities are not reliable. This is illustrated both by figure \ref{fig:scatter-subplot2}, where the diagonal dashed line shows where type 0 and type 1 clusters should be separated, and figure \ref{fig:true-prob-subplot2}, where we can see that the classification probabilities are far from calibrated. This is due to the fact that we use a uncorrelated Gaussian model for the noise, despite the fact this model is wrong.
~\\
\begin{minipage}{0.99\columnwidth}
\vspace{10pt}
\centering
\includegraphics[width=\columnwidth]{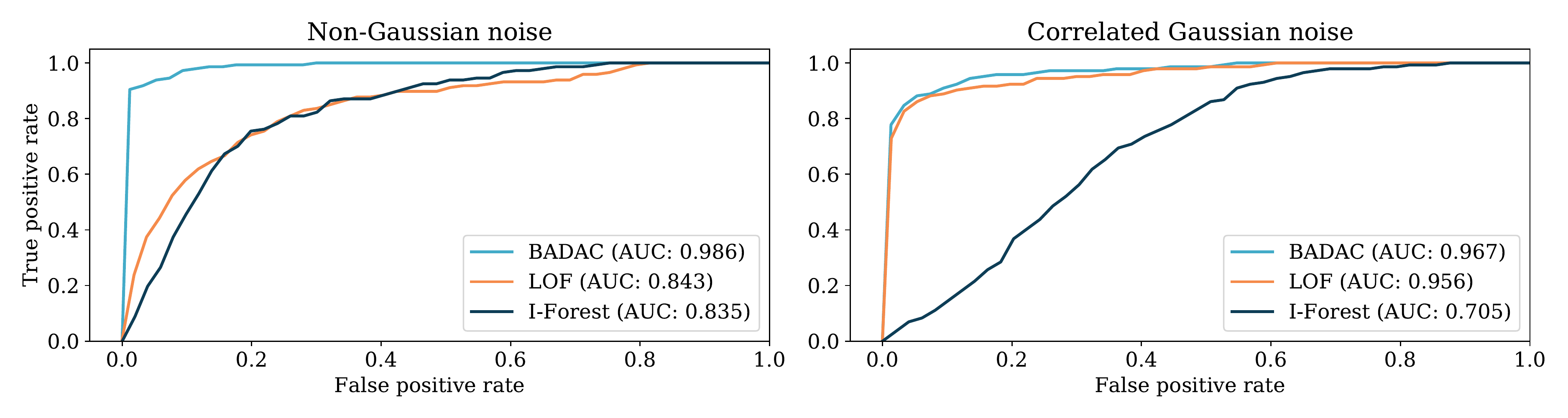}
\captionof{figure}{ROC curves for \mlfeb{anomaly detection with} BADAC, LOF and IsolationForest on the dataset with non-Gaussian error (left pane), and the dataset with correlated Gaussian error (right pane). BADAC performs best in both cases under the AUC metric (values shown in the legend).}
\label{fig:roc-subplot2}
\end{minipage}

We show the ROC curves for BADAC as well as LOF and IsolationForest in figure \ref{fig:roc-subplot2} in order to gauge performance in anomaly detection. In these two cases, it is surprising BADAC performs best, since we don't correctly model the noise. Random forests however achieves a higher accuracy in classification in these two cases. A summary of algorithm performance from all algorithms on all the datasets we consider in both anomaly detection and classification is shown in tables \ref{tab:result-summary} and \ref{tab:result-summary-classification}.

\begin{table}[!h]
\caption{Results for anomaly detection only: Non-Gaussian and correlated Gaussian noise. BADAC shows the best performance in both experiments, showing some robustness to incorrectly choosing the model of the noise. In the non-Gaussian case both IsolationForest and LOF perform poorly in terms of MCC and RWS due to the wide tails allowing for large noise fluctuations.}
\label{tab:result-summary2}
\centering
\begin{tabular}{l c c c c c c c c c}
\hline
 & \multicolumn{3}{c}{BADAC} & \multicolumn{3}{c}{IsolationForest} & \multicolumn{3}{c}{LOF} \\
\hline
 & MCC & AUC & RWS & MCC & AUC & RWS & MCC & AUC & RWS \\
\hline
Non-Gauss. & {\bf 0.84} & {\bf 0.99} & {\bf 0.96} & 0.06 & 0.84 & 0.10 & 0.16 & 0.84 & 0.18 \\
Corr. noise & {\bf 0.68} & {\bf 0.97} & {\bf 0.84} & 0.01 & 0.70 & 0.03 & 0.61 & 0.96 & 0.76 \\
\hline
\end{tabular}
\end{table}

\begin{table}[!h]
\caption{We compare BADAC average accuracy for {\em classification} of all classes to that of Random forests. BADAC performs reasonably in the case of non-Gaussian noise but fairly poorly on the correlated noise case, due to the incorrect model assumption in BADAC, while random forests is more robust as it can learn a model from the training data, while BADAC insists on interpreting the fluctuations as coming from an uncorrelated Gaussian distribution. This relatively poor performance of BADAC can be rectified by using, or learning, the right noise model. }
\label{tab:result-summary-classification2}
\centering
\begin{tabular}{p{7.5cm} c c}
\hline
 & BADAC & Random forests \\
\hline
Non-Gaussian noise & 97.71 & {\bf 98.14} \\
Correlated Gaussian noise & 68.88 & {\bf 96.72} \\
\hline
\end{tabular}
\end{table}

\section{Extensions}\label{sec:extensions}

\subsection{Learning Subclasses}

The BADAC algorithm we have presented can classify and identify anomalies. It can also add new classes as needed when run in online learning mode, as discussed in section \ref{sec:online}. The following \mlfeb{example} illustrates how BADAC can potentially identify subclasses of existing classes. But first, what do we mean by a subclass? Here a subclass corresponds to a large intrinsic variation in a class inconsistent with measurement errors. 

In equation \ref{eq:maineq} we explicitly allowed each instance of a class $\tau$ to have its own true value, $y_{t,j}^i$, for each feature $j$. However a very homogeneous class will not require this flexibility, and will only require one latent variable for each feature $j$. Hence, we can define the number of subclasses to be the number of different latent variables per feature $j$, required by the data in class $\tau$. 

Let us define a hierarchy of models, $\cal{M_\alpha}$ where $\alpha$ is the number of subclasses of class $\tau$ and also the number of true latent variables per feature. Using all the data from class $\tau$ we infer the latent variables $\theta_{ij} \equiv y_{t,j}^i$. We can then select the preferred number of subclasses $\alpha$ by maximising the Bayesian evidence after marginalising over $\theta_{ij}$. The subtlety is that in the models with more than one subclass, we do not know, {\em a priori}, which subclass an instance may belong to. We solve this by introducing new latent parameters for the subclass which allows each instance to belong to any of the subclasses and then writing the likelihood as a mixture model as was done in \citep{beams1,beams2,newling2012,knights2013}. Finally we marginalise over these subclass labels to compute the model evidence and find the preferred number of subclasses. 

\subsection{Dealing with Missing Data}

In our discussion so far we have assumed the idealised case that we have data at the same points for all training and test data. This is clearly unrealistic and an important limitation. How do we deal with missing data? 

There are two approaches. The first, more conservative, approach is to sample from the prior distribution with the error given by the prior  distribution for that class. If the data is missing from test data, the missing data can be sampled as above, but in each case we use the prior for the class that it is being compared against. 

The second approach is to use some form of interpolation. A natural approach is to use Gaussian processes, since these give both an expected value and Gaussian error at the missing data. Gaussian processes need a covariance function which encodes how rapidly the underlying class varies. As a result each class will have their own Gaussian process and covariance function which should be learned from the training data. Test data should then be compared to training classes using the appropriate Gaussian process for each of the training classes.

\subsection{Template construction and speeding up BADAC}\label{sec:templates}

As shown in table \ref{tab:time} the full BADAC calculation is much slower than other classification or anomaly detection algorithms. This stems from the pairwise comparison of all data in the test dataset with all data in the training set, something which becomes computationally infeasible for very large amounts of training data. 

Fortunately in the limit of large training data we can expect to sample the class distribution well and therefore we can instead create a single template for each class (or as an intermediate step, each sub-class), which will dramatically speed up BADAC, though at the cost of having a non-Gaussian spread in general.  

How should we compute class or sub-class templates? An elegant solution is to fit a single Gaussian process to the data of each class \citep{GP_bayes}. This has the advantage of automatically dealing with any missing data, but will not deal with non-Gaussian or multi-model intra-class variability. To get around this limitation one could use a Kernel Density Estimate summed over the training data in each class at each value of the independent variable (or in bins). However, since this is a still a sum over all the training data examples it will be slow. To speed it up we need some approximation to the KDE sum. 

Probably the simplest approximation - which also preserves the Gaussian distribution - is to use the inverse-variance estimator $\hat{y}$ with standard deviation $\hat{\sigma}$: 
\begin{eqnarray}
\hat{y} &&= \hat{\sigma}^2 \sum_i y_i/\sigma_i^2 \\
\hat{\sigma}^2 &&= \left(\sum_i 1/\sigma_i^2\right)^{-1}
\end{eqnarray}
If the intraclass variability is highly non-Gaussian then it would be better to fit a more appropriate low-dimensional distribution to describe this to create the template. We then effectively reduce to the formalism described in section \ref{parametric_BADAC}.

\subsection{Intraclass Variability}

In the formulation presented earlier we assumed that the variability in the observed data for a given class was given by the measurement errors on the observed data, i.e. that the intraclass variability was small. If this is not the case one can build more complex models for the intra-class variability. The simplest is to fit for a global standard deviation, $\sigma_*$, at training for each class (for example by using a validation subset of the training data). The intra-class variability model can be made arbitrarily complex and the Bayesian evidence could be used to select the best model.  

\subsection{Calibration and Zero-point Issues}

In applying BADAC to real examples there may be systematic differences in the data between test and training. This could, for example, be because the data comes from different instruments or is taken under different conditions. As an example, consider applying BADAC to images where there may be large-scale calibration differences across the images. How can one deal with such effects which will invalidate the use of the simple versions of the BADAC formalism presented earlier, along with most anomaly and classification algorithms?  

In the spirit of the Bayesian approach, one way to deal with such large-scale artefacts is to model their effects and introduce nuisance parameters $\varphi$, with their own prior distributions $P(\varphi)$, which are then marginalised over before classification or anomaly ranking. Intuitively this means that the algorithm will exploit the freedom implicit in the calibration model to try to fit each test curve to the training data and will only highlight as outliers those data which are poor fits no matter the calibration freedom.

A related problem is the issue of zero-points, which occurs if the examples in training and test data are not all aligned on the $x$-axis. This is common when working with time series data. In principle this can be dealt with in a similar way, by allowing each data example to have an extra translation parameter which allows one to shift all points in the example left or right. One must then marginalise over this nuisance parameter when doing the fits. 

Depending on the exact nature of the data these translation parameters may be well-constrained. For example, one may be able to align all examples approximately, in which case one can put priors on the translation parameters. However, the zero-point issue does raise significant complications. For each pair in the training and test sets one should in principle allow a translation parameter. This leads to $n \times M$ new nuisance parameters where $n,M$ are the number of training and test set examples respectively. Unless the marginalisation can be performed analytically this will typically be prohibitively expensive. 

A cheaper alternative is to pre-align all the training data by class. Now there is only one translation nuisance parameter per class and per instance in the test set. However,  the alignment of the training data will not be perfect in general. This can be handled by adding an $x$-error bar to each data point in the training data, corresponding to small errors in the alignment of the data. These $x$-errors are perfectly correlated however (since the translation affects all data in the same way) and the BADAC formalism would need to be extended to account for such correlations, as done in e.g. \cite{xfisher,zbeams}.

\section{Conclusions}

We have presented a novel statistically robust joint anomaly detection and classification method, Bayesian Anomaly Detection And Classification (BADAC), that is designed to take advantage of any knowledge of the underlying noise distribution in the training and test data. Although we perform tests for the case of Gaussian distributed data, our formalism is general. 

\mlfeb{Using simulated one-dimensional data, we test the classification and anomaly detection capabilities of BADAC. We make use of several metrics, including our novel Rank-Weighted-Score that rewards algorithms for ranking more anomalous objects above those that have been commonly seen. We find that in the case where the correct noise model is known BADAC outperforms random forests at classification and both IsolationForest and local outlier factor (LOF) at anomaly detection, due to its ability to correctly exploit uncertainty information. In the case of compact anomalies, which could emulate noisy spikes in data, we find that BADAC's performance is comparable to LOF and superior to IsolationForest.} We demonstrate how BADAC produces calibrated classification probabilities, \mlfeb{which is crucial if a machine learning algorithm is to be incorporated into a precise, scientific analysis pipeline}.

\mlfeb{We performed tests to investigate the degradation of performance if the assumptions of BADAC are violated. Interestingly, we find BADAC still outperforms the other anomaly detection algorithms in the presence of non-Gaussian and correlated noise. However we find its classification performance degrades, especially in the correlated case. We also note that with an incorrect noise model, the probabilities of BADAC are no longer guaranteed to be calibrated. However, if the structure of the noise is known, the correct noise model can be incorporated into the BADAC likelihood.} 

While BADAC provides excellent performance by exploiting the extra information about the underlying noise distributions, the computational limitations discussed in section \ref{sec:computational} mean that it does not scale well to large training datasets. In this case one must either use prototype templates to represent the classes (e.g. through Gaussian processes) or parameterise the data, to speed up classification and anomaly detection with BADAC.

\mlfeb{We find ourselves in an era of exponentially increasing data volume, driving the need for machine learning algorithms. However in the physical sciences there is equal need for accurate propagation of uncertainties from all parts of an analysis pipeline, including any machine learning algorithms. With its statistically principled approach to both classification and anomaly detection, BADAC is able to provide believable and interpretable probabilities in the presence of measurement uncertainties, as required by high precision scientific analysis.}

\begin{acknowledgements}
We thank Alireza Vafaei Sadr, Martin Kunz and Boris Leistedt for discussions and comments. We acknowledge the financial assistance of the National Research Foundation (NRF). Opinions expressed and conclusions arrived at, are those of the authors and are not necessarily to be attributed to the NRF. This work is partially supported by the European Research Council under the European Community's Seventh Framework Programme (FP7/2007-2013)/ERC grant agreement no 306478-CosmicDawn.
\end{acknowledgements}

\bibliographystyle{spbasic}      % basic style, author-year citations
\bibliography{references}   % name your BibTeX data base

\pagebreak

\appendix

\section{Detailed derivation of BADAC}\label{sec:detailed-deriv}

In this section we go through the details of the formalism we discussed in section \ref{sec:BHM}. We begin with the simple case of one training and test datapoint, and then add $n$ training points. We finish by generalising the formalism to objects with $m$ datapoints each.

\subsection{Modelling one training point}\label{sec:BHM1}

Our goal is to determine the probability of an object $d$ belonging to class $\tau$. This is calculated assuming that we have an observation of another object $y_o$, which we know is from class $\tau$. The probability of $d$ belonging to class $\tau$, given $y_o$ can hence be written as $P(\tau|d,y_o)$. By implementing Bayes' theorem, we can arrive at equation \ref{eq:bayesA}:
\begin{equation}\label{eq:bayesA}
P(\tau|d,y_o) \propto P(d,y_o|\tau) P(\tau)
\end{equation}

\begin{figure}[h]
\includegraphics[width=8cm]{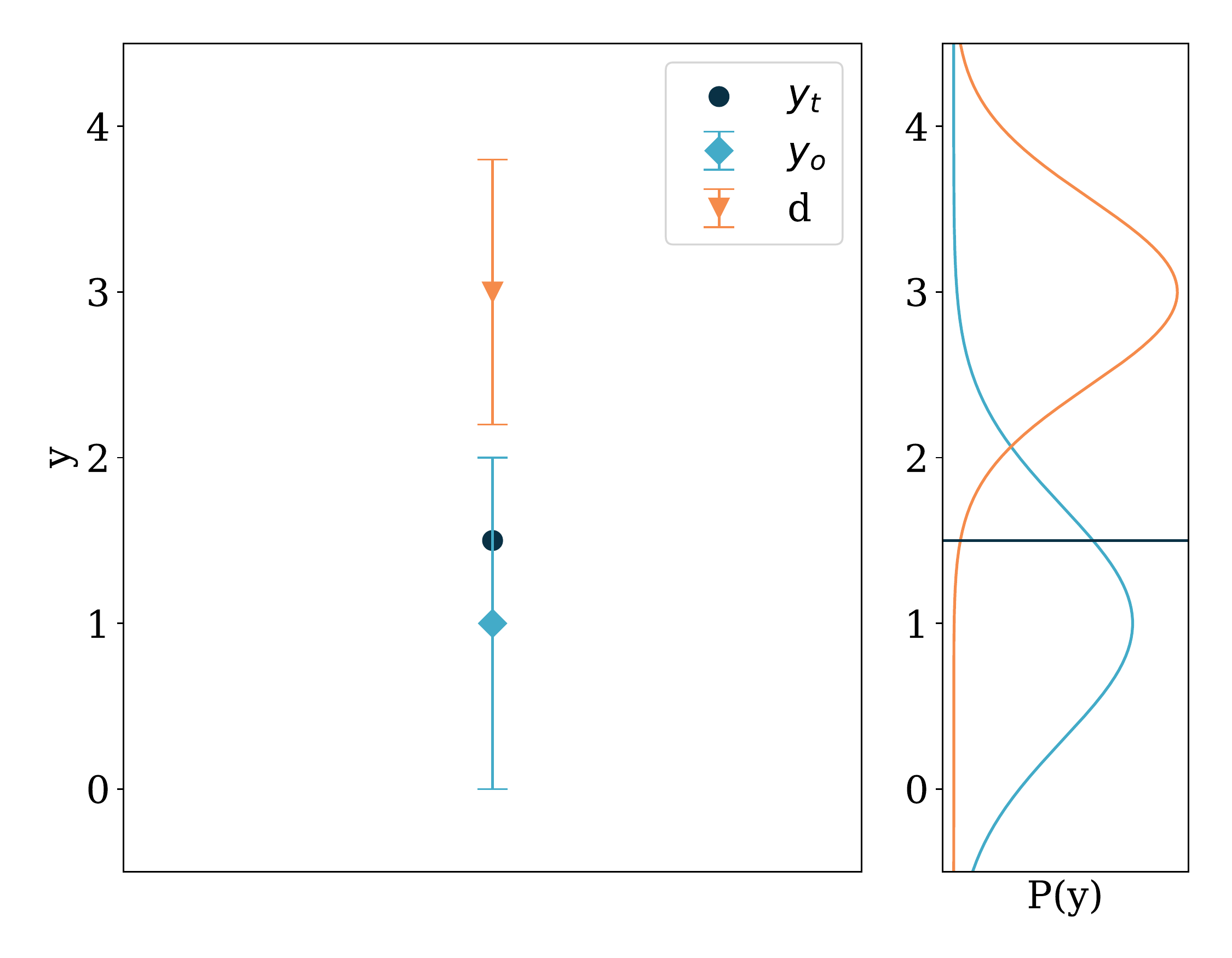}
\centering
\caption{\small{Plot of the schematic representation of this Bayesian hierarchical model with a single datapoint. This representation is shown in the left pane, and the probability distributions associated with the uncertainty on the data are shown in the right pane. The orange triangular marker for $d$ shows the new test point we would like to consider. The light blue diamond marker for $y_o$ shows the training datum of known class type. The black dot represents $y_t$, a latent parameter that we introduce that can be thought of as the ``true value'' for $y_o$, if we were able to know $y_o$ exactly. This is the parameter we wish to marginalise over.}}
\label{fig:app:schem}
\end{figure}

If we now assume that we have a measurement error associated with both $d$ and $y_o$, we cannot expand equation \ref{eq:bayesA} to calculate the probability as is. To solve this we introduce a latent parameter $y_t$, which we wish to marginalise over. Here $y_t$ is the ``true" value of $y_o$, as though we were able to know the correct value of $y_o$ exactly. By application of the product rule, we obtain the result shown in equation \ref{eqn:product-A}.
\begin{align}
P(\tau|d,y_o) &\propto P(\tau) \int dy_t P(d,y_o,y_t|\tau)\\
P(\tau|d,y_o) &\propto P(\tau) \int dy_t P(d,y_o|y_t,\tau) P(y_t|\tau)\\
P(\tau|d,y_o) &\propto P(\tau) \int dy_t P(d|y_o,y_t,\tau) P(y_o|y_t,\tau) P(y_t|\tau) \label{eqn:product-A}
\end{align}
We illustrate this hierarchical model graphically in figure \ref{fig:app:schem}. If we assume $d$ and $y_o$ are statistically independent of one another, then we can drop the dependence on $y_o$ from terms that depend on $y_t$ as well i.e.: $P(d|y_o,y_t,\tau) = P(d|y_t,\tau)$. We can do this since we assume that $y_t$ represents $y_o$ exactly. We then end up with the result shown in equation \ref{eq:app:bhm1}. 
\begin{equation}\label{eq:app:bhm1}
P(\tau|d,y_o) \propto P(\tau) \int dy_t P(d|y_t,\tau) P(y_o|y_t,\tau) P(y_t|\tau)
\end{equation}
We assume the measurement errors associated with $d$ and $y_o$ are Gaussian, which makes the probability terms  $P(d|y_t,\tau)$ and $P(y_o|y_t,\tau)$ Gaussian likelihoods. We further assume a flat prior on the distribution of true values of $y_o$, $P(y_t|\tau)$. These assumptions allow us to perform the integration analytically. We can now generalise this method to work for $n$ training/known objects. 
%[ML] I don't think this is necessary 
%We do this in section \ref{sec:BHMn}.

\subsection{Modelling $n$ training instances}\label{sec:BHMn}

Here we begin with Bayes' theorem in the same manner as before. The class probability is now dependent on the unknown object $d$, as well as $n$ known objects $y_o^i$ where $i \in [1,n]$:
\begin{equation}\label{eq:bayes_n}
P(\tau|d,y_o^1,...,y_o^n) \propto P(d,y_o^1,...,y_o^n|\tau) P(\tau)
\end{equation}

Now we introduce $n$ latent variables $y_t^i$, with $i\in [1,n]$, such that $y_t^i,...,y_t^n$ are the true values corresponding to each object $y_o^i$. By marginalising over these latent variables, we arrive at equation \ref{eq:latent1}.
\begin{equation}\label{eq:latent1}
P(\tau|d,y_o^1,...,y_o^n) \propto P(\tau) \int dy_t^1...dy_t^n P(d,y_o^1,...,y_o^n,y_t^1,...,y_t^n|\tau)
\end{equation}

We can apply the product rule, where we group all variables $y_t^i$ together. We can do this since $P(A,B,C) = P(A,B|C) P(C) = P(A|B,C) P(B|C) P(C) = P(A|B,C) P(B,C)$.
\begin{equation}
P(\tau|d,y_o^1,...,y_o^n) \propto P(\tau) \int dy_t^1...dy_t^n P(d,y_o^1,...,y_o^n|y_t^1,...,y_t^n,\tau) P(y_t^1,...,y_t^n|\tau)
\end{equation}

Now we apply product rule again. We also skip a step we showed in section \ref{sec:BHM1}, where we drop the dependence of $y_o^1,...,y_o^n$ on $P(d|y_o^1,...,y_o^n,y_t^1,...,y_t^n,\tau)$. This does not apply if observations of different objects are correlated, but for clarity, we ignore this possible source of correlation. 
\begin{equation}\label{eq:latent2}
P(\tau|d,y_o^1,...,y_o^n) \propto P(\tau) \int dy_t^1...dy_t^n P(d|y_t^1,...,y_t^n,\tau) P(y_o^1,...,y_o^n|y_t^1,...,y_t^n,\tau) P(y_t^1,...,y_t^n|\tau)
\end{equation}

In addition to assuming the objects $y_o^i$ are statistically independent of one another, we assume they are independent of the true values $y_t^j$, where $j \neq i$ as well. This means that $P(y_o^1,...,y_o^n|y_t^1,...,y_t^n,\tau) = P(y_o^1|y_t^1,\tau) ... P(y_o^n|y_t^n,\tau)$. We can therefore express equation \ref{eq:latent2} as follows:
\begin{equation}\label{eq:latent3}
P(\tau|d,y_o^1,...,y_o^n) \propto P(\tau) \int dy_t^1...dy_t^n P(d|y_t^1,...,y_t^n, \tau) \prod_{i=1}^n P(y_o^i|y_t^i,\tau) \prod_{i=1}^n P(y_t^i|\tau)
\end{equation}

Now we need to find a way to calculate the value of $P(d|y_t^1,...,y_t^n, \tau)$. The probability of observing object $d$ given one $y_t$ is a simple likelihood. We use a kernel density estimate for the probability of observing $d$ given a number of variables $y_t^i$. This means that $P(d|y_t^1,...,y_t^n, \tau) \equiv \frac{1}{n} \sum_{i=1}^n P(d|y_t^i,\tau)$. Now we have:
\begin{equation}\label{eq:latent4}
P(\tau|d,y_o^1,...,y_o^n) \propto P(\tau) \int dy_t^1...dy_t^n \left[\dfrac{1}{n} \sum_{i=1}^n P(d|y_t^i,\tau) \right] \prod_{i=1}^n P(y_o^i|y_t^i,\tau) \prod_{i=1}^n P(y_t^i|\tau)
\end{equation}

$P(y_t^i|\tau)$ is simply a prior on the true $y_o^i$ measurements. Since we don't aim to place any constraints on the intraclass variability of class $\tau$, we choose a flat prior here ($P(y_t^i|\tau)=1$). It is important to note that the choice of an improper prior on hierarchical parameters can introduce biases on parameter estimates (\cite{gull1989} and \cite{gelman2006}). One could easily select a more informative prior if there was additional information about the data available. For ease, the prior should be analytically integrable with the form of the probabilities $P(d|y_t^i,\tau) P(y_o^i|y_t^i,\tau)$. It would however be possible to resort to numerical integration if this were not the case. Here we assume uncorrelated Gaussian measurement error in the determination of the values of $P(d|y_t^i,\tau)$ and $P(y_o^i|y_t^i,\tau)$, so equation \ref{eq:latent4} can be expressed as the following:
\begin{equation}\label{eq:latent5}
P(\tau|d,y_o^1,...,y_o^n) \propto \int_{-\infty}^{\infty} dy_t^1...dy_t^n \left[ \frac{1}{n}\sum^n_{i=1} \frac{1}{\sqrt{2\pi}\sigma_d} \exp\left( -\frac{1}{2}\left(\dfrac{d-y_t^i}{\sigma_d}\right)^2 \right) \right] \prod_{i=1}^n \frac{1}{\sqrt{2\pi}\sigma_{y_o^i}} \exp\left(-\dfrac{1}{2}\left(\dfrac{y_o^i-y_t^i}{\sigma_{y_o^i}}\right)^2\right),
\end{equation}
where $\sigma_d$ is the uncertainty on $d$ and $\sigma_{y_o^i}$ is that on the $i$'th training point $y_o$. Note that we can switch the sum and integral, and then evaluate the integral of the product of $n+1$ Gaussian likelihoods. However, when we evaluate this integral, it simplifies in the case where the set of $y_t^i$ are uncorrelated. The simplification also means we can integrate over $dy_t^i$ instead of $dy_t^1...dy_t^n$.
\begin{equation}\label{eq:latent6}
P(\tau|d,y_o^1,...,y_o^n) \propto \frac{1}{n}\sum^n_{i=1} \int_{-\infty}^{\infty} dy_t^i \frac{1}{\sqrt{2\pi}\sigma_d} \exp\left(-\dfrac{1}{2}\left(\dfrac{d-y_t^i}{\sigma_d}\right)^2\right) \frac{1}{\sqrt{2\pi}\sigma_{y_o^i}} \exp\left( -\dfrac{1}{2}\left(\dfrac{y_o^i-y_t^i}{\sigma_{y_o^i}}\right)^2 \right)
\end{equation}

Now it is easy to complete the square and do the integration analytically. In equation \ref{eq:bhmfinal} we simply state the result:
\begin{equation}\label{eq:bhmfinal}
P(\tau|d,y_o^1,...,y_o^n) \propto \frac{1}{n}\sum^n_{i=1} P(\tau_i) (2\pi \sigma_d \sigma_{y_o^i})^{-1} \left[ \dfrac{\pi}{\frac{1}{2}(\Gamma_d + \Gamma_i)} \right]^{1/2} \exp\left(-\frac{1}{2}\left(\Gamma_d d^2 + \Gamma_i {y_o^i}^2 -\dfrac{(\Gamma_d d + \Gamma_i y_o^i)}{\Gamma_d +\Gamma_i}\right)\right)
\end{equation}
where $\Gamma_d = \dfrac{1}{\sigma_d^2}$ and $\Gamma_i = \dfrac{1}{{\sigma_{y_o^i}^2}}$.

Equation \ref{eq:bhmfinal} only handles objects with a single datapoint. For objects with multiple datapoints, we introduce $P(\tau|\{d_j\},\{y_{o,j}^1\},...,\{y_{o,j}^n\})$, where it is a product of the probabilities for the single datapoints assuming they are uncorrelated. This means we have:
\begin{multline}\label{eq:bhmfinalm}
P(\tau|\{d_j\},\{y_{o,j}^1\},...,\{y_{o,j}^n\}) \propto \frac{1}{n}\sum^n_{i=1} P(\tau_i) \prod_{j=1}^m (2\pi \sigma_{d_j} \sigma_{y_{o,j}^i})^{-1} \\ 
\times\ \left[ \dfrac{\pi}{\frac{1}{2}(\Gamma_d + \Gamma_i)} \right]^{1/2} \exp\left(-\frac{1}{2}\left(\Gamma_d d_j^2 + \Gamma_i {y_{o,j}^i}^2 -\dfrac{(\Gamma_d d + \Gamma_i y_{o,j}^i)}{\Gamma_d + \Gamma_i}\right)\right)
\end{multline}
where $\Gamma_d = \dfrac{1}{\sigma_{d,j}^2}$ and $\Gamma_i = \dfrac{1}{{\sigma_{y_{o,j}^i}^2}}$, and $\{d_j\}$ and $\{y_{o,j}^i\}$ are all sets over the index $j$.

We use equation \ref{eq:bhmfinalm} in our evaluation in section \ref{sec:results}.

\section{Correlated data}\label{sec:corr}

It is possible to take correlated data into account in the formalism we've developed, if it is known how the data are correlated. Taking into account correlated data allows for better classification accuracy. Here we will talk about two types of correlation. Firstly, correlations may exist between different features in the same instance. Here we'll refer to this as intra-instance correlation. Then we have correlations between different instances, which we refer to as inter-instance correlation. The following subsections show where accounting for these correlations would enter the formalism we developed in sections \ref{sec:BHM1} and \ref{sec:BHMn}.

\subsection{Intra-instance correlation}\label{sec:intra-corr}

Intra-instance correlations often arise in science. These are also the type of correlations we assess in section \ref{sec:results-nongauss-corr}, albeit without correctly accounting for the correlated nature of the data in order to test the robustness of our formalism. Here we show how one would correctly account for correlated Gaussian noise using BADAC.

Continuing from equation \ref{eq:latent4}, we get the following result if we assume that $P(d|y_t^i,\tau)$, $P(y_o^i|y_t^i,\tau)$ and $P(y_t^i|\tau)$ are normalised Gaussian distributions, and statistically independent of one another (we explore what happens when this is not the case in section \ref{sec:inter-corr}):
\begin{align}
P(\tau|d,y_o^1,...,y_o^n) &\propto P(\tau) \int dy_t^1...dy_t^n \left[\dfrac{1}{n} \sum_{i=1}^n P(d|y_t^i,\tau) \right] \prod_{i=1}^n P(y_o^i|y_t^i,\tau) \prod_{i=1}^n P(y_t^i|\tau) \\
P(\tau|d,y_o^1,...,y_o^n) &\propto P(\tau) \dfrac{1}{n}\sum_{i=1}^n \int dy_t^i P(d|y_t^i,\tau) P(y_o^i|y_t^i,\tau) P(y_t^i|\tau)
\end{align}

Evaluating $P(d|y_t^i,\tau)$ when there is correlated Gaussian noise is conceptually simple through the introduction of a covariance matrix $C_d$.In practice, estimating $C_d$ can be difficult. One approach is to parametrise the covariance matrix and marginalise over these nuisance parameters. Assuming we know the covariance matrix, we can evaluate $P(d|y_t^i,\tau)$ for the $i$-th instance as follows:
\begin{equation}
P(d|y_t^i,\tau) =  (2\pi \det|C_d|)^{-\frac{1}{2}} \exp \left( -\frac{1}{2}{\Delta_d^i}^T {C_d}^{-1} \Delta_d^i \right)
\end{equation}
where $\Delta_d^i = (d_1-y_{t,1}^i, \hdots, d_m-y_{t,m}^i)^T$, and $C_d$ is the covariance matrix of the observed instance $d$, where $d$ has $m$ features. In the same way, we can evaluate $P(y_o^i|y_t^i,\tau)$ for the $i$-th instance as follows:
\begin{equation}
P(y_o^i|y_t^i,\tau) = (2\pi \det|C_{y_o^i}|)^{-\frac{1}{2}} \exp \left( -\frac{1}{2}{\Delta_{y_o^i}}^T {C_{y_o^i}}^{-1} \Delta_{y_o^i} \right)
\end{equation}

$\Delta_y^i = (y_{o,1}^i-y_{t,1}^i, \hdots, y_{o,m}^i-y_{t,m}^i)^T$ and $C_y^i$ are the $n$ intra-instance covariance matrices of the observed training instances $y_o^i$.

\subsection{Inter-instance correlation}\label{sec:inter-corr}

A possible source of correlations is that the training curves $y_o^i$ could be statistically dependent on one another. This would mean that for a measurement of an object $y_o^i$, another object $y_o^j$ might be scattered in a correlated or anti-correlated way relative to $y_o^i$. Here we will call this type of correlation inter-object correlation. Possible causes of this type of correlation in a physical system would be biases introduced by measurement equipment, or observing/environmental conditions. What this would mean for the formalism in sections \ref{sec:BHM1} and \ref{sec:BHMn}, is that we would not be able to assume $P(d|y_o^i,...,y_o^n, y_t,...,y_t^n,\tau) = P(d|y_t^i,...,y_t^n,\tau)$. It is not trivial to evaluate $P(d|y_o^i,...,y_o^n,y_t^i,...,y_t^n,\tau)$ sensibly, which is why we assume the dependence of $d$ on $y_o^i,...,y_o^n$ is negligible in equation \ref{eq:bhm1}. 

Inter-object correlations also enter into our formalism in equation \ref{eq:latent3}. Here we assumed that $P(y_o^1,...,y_o^n|y_t^1,...,y_t^n,\tau) = P(y_o^1|y_t^1,\tau) ... P(y_o^n|y_t^n,\tau)$. If one were interested in taking these correlations into account, the correct expansion would be:

\begin{align}
P(y_o^i,...,y_o^n|y_t^i,...,y_t^n,\tau) &= P(y_o^1,...,y_o^n|y_t^1,...,y_t^n,\tau)\\
P(y_o^i,...,y_o^n|y_t^i,...,y_t^n,\tau) &= P(y_o^n|y_t^n,\tau) \prod_{i=1}^{n-1} P(y_o^i|y_o^{i+1},...,y_o^n,y_t^i,...,y_t^n,\tau)\label{eq:corr1}
\end{align}

If we assume a Gaussian form for the likelihood as in equation \ref{eq:latent5}, and further assume no intra-instance correlations (no $x-x'$ correlations), then equation \ref{eq:corr1} simplifies to:

\begin{equation}\label{eq:corr2}
P(y_o^i,...,y_o^n|y_t^i,...,y_t^n,\tau) = (2\pi \det|C|)^{-\frac{1}{2}} \exp \left(-\frac{1}{2}\left( \Delta^T C^{-1} \Delta \right)\right)
\end{equation}
where $\Delta = (y_o^1-y_t^1, \hdots, y_o^n-y_t^n)^T$
and $C$ is the covariance matrix.

This type of correlation could arise for example, if multiple training data are observed with the same observing factors, leading to correlated errors.

\section{Metrics for Algorithm Evaluation}\label{sec:metrics}

In this section we outline the metrics we use in section \ref{sec:algo-perform} to quantify algorithm performance for both classification and anomaly detection. The choice of a metric is important since inappropriate metrics can give very misleading results. In our case we want metrics that are insensitive to class imbalance (since anomalies are assumed to be rare). 

While one of the metrics considered, the AUC (section \ref{sec:auc}), uses the probability of belonging to a particular class, the other metrics discussed require a strict classification. In all cases, we take the class with the highest probability to be the algorithm's classification.

\subsection{Area Under the Curve}
\label{sec:auc}

The Area Under the Curve (AUC) metric is suitable for gauging performance in any binary classification problem. It is found by calculating the area under the Receiver Operating Characteristic (ROC) curve for the class of interest. The ROC curve is found by plotting the True Positive Rate (TPR) against the False Positive Rate (FPR). In order to do this, some sort of probability of belonging to the class of interest is required for the objects being classified.

The AUC has a possible range of $[0,1]$, where $0$ represents all true objects belonging to the class of interest being given lower probabilities than other objects, and $1$ represents all true objects belonging to the class of interest being given higher probabilities than other objects. An algorithm performs perfectly under this metric with a score of $1$, and the worst possible performance is $0$. Random guessing for classification would on average yield an AUC score of $0.5$ for either balanced or unbalanced data.

Note that the AUC of the ROC curve is known to be problematic for imbalanced data (of which anomaly detection is the extreme case) \cite{auc}. As a result we also consider the MCC and RWS scores.

\subsection{Matthews Correlation Coefficient}

The Matthews Correlation Coefficient (MCC) \citep{matthews1975} is another metric for assessing the performance of machine learning algorithms on binary classification problems. It is useful for measuring algorithm performance on datasets with unbalanced classes. For example, an algorithm would score $99\%$ accuracy on a dataset with 99 inliers and one outlier if it predicted only inliers. The same algorithm would get an MCC score of $0$, where $-1$ is the worst possible score, and $1$ the best.

The MCC is defined as:
\begin{equation}
MCC = \dfrac{TP\times TN - FP\times FN}{\sqrt{(TP+FP)(TP+FN)(TN+FP)(TN+FN)}}
\end{equation}
where TP is the number of true positives, TN the number of true negatives, FP the number of false positives and FN the number of false negatives in a test dataset.

In this section we only use the MCC score to measure performance in anomaly detection, not classification, and the positive class refers to the anomaly class.

\subsection{Accuracy}

We use the overall accuracy in computing classification performances only (not anomaly detection) in section \ref{sec:algo-perform}. This is the number of correctly classified objects divided by the total number of objects.

\end{document}